\documentclass{article}

\usepackage{microtype}
\usepackage{graphicx}
\usepackage{booktabs} %

\usepackage{hyperref}

\usepackage{inconsolata}
\usepackage{tabulary, ragged2e}
\usepackage{xcolor}
\usepackage{soul}
\usepackage{tablefootnote}

\usepackage{hyperref}
\usepackage{url}

\usepackage{caption}
\usepackage{subcaption}
\usepackage[inkscapelatex=false]{svg}

\usepackage{graphicx}
\usepackage{siunitx}

\newcommand{\gptfour}{\texttt{GPT-4}}
\newcommand{\gptturbo}{\texttt{InstructGPT} (turbo)}
\newcommand{\gptthreex}{\texttt{ChatGPT-3.5}}
\newcommand{\gptthree}{\texttt{InstructGPT} (text-davinci-003)}
\newcommand{\gpttwo}{\texttt{InstructGPT} (text-davinci-002)}
\newcommand{\palm}{\texttt{PaLM2 (text-bison)}}
\newcommand{\chatgpt}{\texttt{ChatGPT}}

\newcommand{\falcon}{\texttt{Falcon-180B-Chat}}

\usepackage[accepted]{style/icml2024}

\usepackage{amsmath}
\usepackage{amssymb}
\usepackage{mathtools}
\usepackage{amsthm}

\usepackage{booktabs}
\usepackage{inconsolata}
\usepackage{tabulary, ragged2e}
\usepackage{subcaption}
\usepackage{xcolor}
\usepackage{soul}
\usepackage{tablefootnote}

\usepackage{hyperref}
\usepackage{url}

\usepackage[noabbrev,capitalize,nameinlink]{cleveref}

\usepackage{caption}
\usepackage{subcaption}
\usepackage[inkscapelatex=false]{svg}
\usepackage{adjustbox}

\newcommand{\ctext}[2]{%
  \begingroup
  \sethlcolor{#1}%
  \hl{#2}%
  \endgroup
}
\definecolor{clipink}{HTML}{FFAABB}
\definecolor{clipyellow}{HTML}{EEDD88}
\definecolor{cliporange}{HTML}{EE8866}
\definecolor{clipear}{HTML}{BBCC33}
\definecolor{clipblue}{HTML}{77AADD}
\definecolor{clip}{HTML}{44BB99}

\newcommand{\refer}{\textsc{ref.}}
\newcommand{\regio}{\textsc{reg.}}
\newcommand{\categ}{\textsc{cat.}}

\theoremstyle{plain}

\theoremstyle{definition}

\theoremstyle{remark}

\icmltitlerunning{Assessing Large Language Models on Climate Information}

\begin{document}

\twocolumn[
\icmltitle{Assessing Large Language Models on Climate Information}

\icmlsetsymbol{equal}{*}

\begin{icmlauthorlist}
\icmlauthor{Jannis Bulian}{equal,gdm}
\icmlauthor{Mike S.\ Sch\"{a}fer}{equal,uzh}
\icmlauthor{Afra Amini}{equal,gdm,eth}
\icmlauthor{Heidi Lam}{equal,goog}
\icmlauthor{Massimiliano Ciaramita}{equal,gdm}
\icmlauthor{Ben Gaiarin}{equal,gdm}
\icmlauthor{Michelle Chen H\"{u}bscher}{equal,gdm}
\icmlauthor{Christian Buck}{equal,gdm}
\icmlauthor{Niels G.\ Mede}{equal,uzh}
\icmlauthor{Markus Leippold}{equal,gdm,uzhf}
\icmlauthor{Nadine Strau{\ss}}{equal,uzh}
\end{icmlauthorlist}

\icmlaffiliation{gdm}{Google DeepMind}
\icmlaffiliation{goog}{Google}
\icmlaffiliation{uzh}{IKMZ - Dept.\ of Communication and Media Research, University of Zurich, Switzerland}
\icmlaffiliation{uzhf}{Dept.\ of Finance, University of Zurich, Switzerland}
\icmlaffiliation{eth}{ETH AI Center, ETH, Zurich, Switzerland}

\icmlcorrespondingauthor{Jannis Bulian}{jbulian@google.com}
\icmlcorrespondingauthor{Mike S.\ Sch\"{a}fer}{m.schaefer@ikmz.uzh.ch}
\icmlcorrespondingauthor{Massimiliano Ciaramita}{massi@google.com}

\icmlkeywords{Machine Learning, ICML}

\vskip 0.3in
]

\printAffiliationsAndNotice{\textsuperscript{*}Authors in random order.}

\begin{abstract}
As Large Language Models (LLMs) rise in popularity, it is necessary to assess their capability in critically relevant domains. We present a comprehensive evaluation framework, grounded in science communication research, to assess LLM responses to questions about climate change. Our framework emphasizes both presentational and epistemological adequacy, offering a fine-grained analysis of LLM generations spanning 8 dimensions and 30 issues. Our evaluation task is a real-world example of a growing number of challenging problems where AI can complement and lift human performance. We introduce a novel protocol for scalable oversight that relies on AI Assistance and raters with relevant education. We evaluate several recent LLMs on a set of diverse climate questions. Our results point to a significant gap between surface and epistemological qualities of LLMs in the realm of climate communication.

\end{abstract}

\section{Introduction}
As concerns around \emph{climate change} intensify~\citep{Poushter-2022, WHO-2021}, more and more people turn to digital media as their primary source of information~\citep{DigNews-2020}. However, in spite of ubiquitous access to information, there remains a considerable gap in climate literacy, exacerbated by the spread of mis- and disinformation~\citep{six-americas-2021}. The challenge of conveying climate data arises also from the nature of scientific communication: science, as an evolving domain, is laden with specialized knowledge, complexity, and inherent uncertainties~\citep{moser-2016-comm}. The digital media landscape, characterized by soaring amounts of AI-generated content~\citep{thompson2024shocking}, limited attention spans and adversarial dynamics, further compounds these challenges~\citep{pearce-2019-some}.

\begin{figure*}[t]
     \centering
     \includegraphics[width=1.0\textwidth]{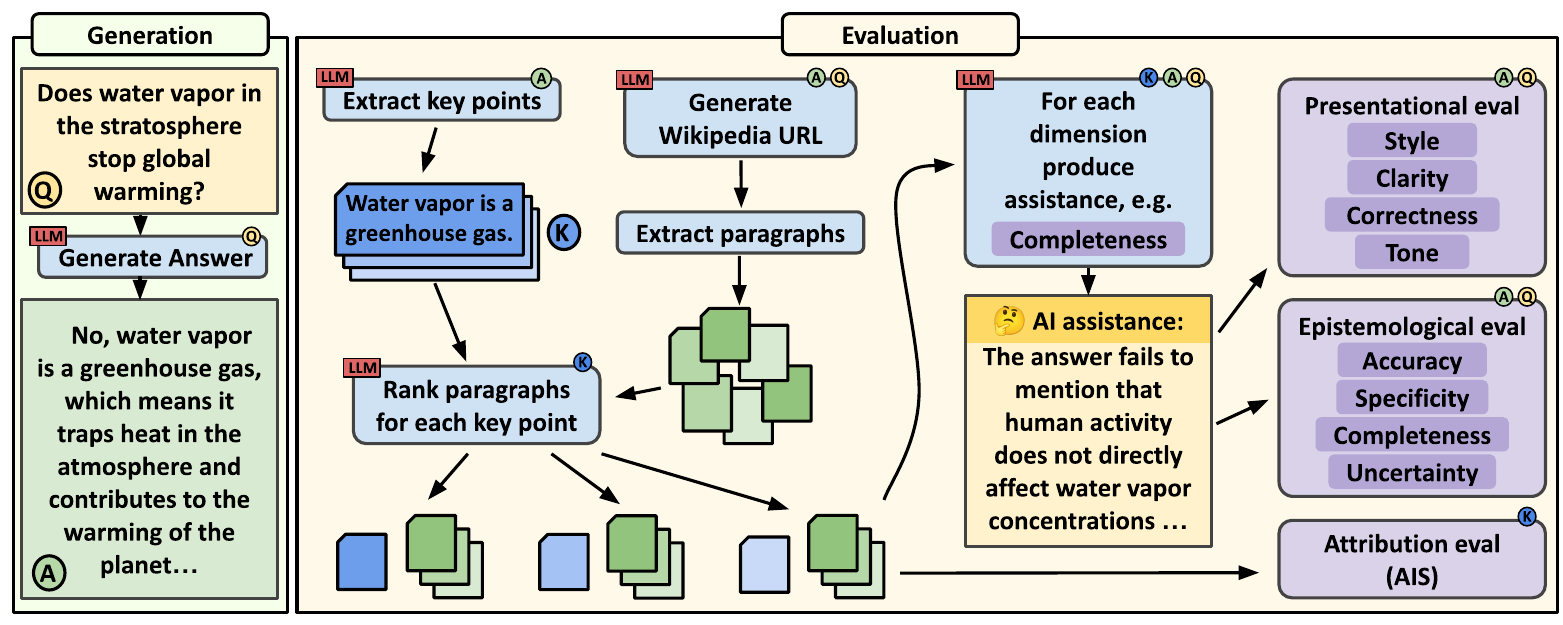}
     \caption{Overview of the evaluation pipeline as described in \cref{sec:experiments}. Starting with a question-answer pair, we use an LLM to extract key points from the answer. We also use the LLM to find a relevant Wikipedia page from which we extract paragraphs. For each key point we rank the paragraphs and keep the top ones. We combine all this information to generate AI assistance for each of our evaluation dimensions. Presentational dimensions are evaluated without the additional paragraphs. This assistance, if available, is presented to our human raters along with the answer. Note that the raters for presentational and epistemological dimension are not shown the key points or retrieved paragraphs. We however use the key points and paragraphs evaluate attribution, cf.~\cref{sec:ais}.}
     \label{fig:sys-design}
\end{figure*}

While AI's promise in addressing global climate challenges is evident through its applications in climate modeling, energy optimization, and disaster management~\citep{CCAI}, its intersection with Natural Language Processing (NLP) is still under-explored. Given recent advancements in LLMs~\citep{gpt-3, palm, openai2023gpt4,geminiteam2023gemini} there is hope that generative AI will also help addressing climate information challenges.
However, using LLMs to address science-related information raises factuality concerns~\citep{ethical-and-social-risks-of-lms}.
Eloquence and advanced dialogue behaviors are trusted by users, even in the absence of trustworthy information \citep{Chiesurin23}. This makes evaluating LLMs difficult.
Research on evaluating systems that may achieve or exceed human abilities, or \emph{scalable oversight}~\citep{Amodei16} is so far mostly theoretical \citep{Irving18, Leike18, Christiano18}, with some recent more practical advances \citep{michael23}. 

We introduce a framework based on
Science Communication research~\citep{oxford-sciencecomm}, to begin evaluating LLMs' responses within the climate change context in a principled way.\footnote{To aid reproducibility, we provide the exact evaluation protocols and all prompts used to generate data.} The evaluation relies on raters with relevant educational background. We assess {\bf presentational} properties such as \emph{style}, \emph{clarity}, linguistic \emph{correctness}, and \emph{tone}. More importantly, we also assess {\bf epistemological} issues: \emph{accuracy}, \emph{specificity}, \emph{completeness}, and \emph{uncertainty}. 
To test the relevance of the evaluation, we run an empirical study on a diverse set of 300 climate change-related questions involving some of the most recent and prominent LLMs. 

Our main findings are as follows: \vspace{-0.4cm}
\begin{itemize}
    \item To increase the recall of detected issues and improve rating quality, it is crucial to introduce scalable oversight protocols that use grounded AI Assistance (cf.\ \cref{fig:sys-design}). However, while AI assistance demonstrably improves rating quality, its influence on raters extends beyond this enhancement. Understanding and mitigating these broader effects remains an open question for future research.
    \item Results suggest that the epistemological quality of responses on climate information of current LLMs is substantially lower than the presentational quality.
    \item We find preliminary evidence that summarizing the evaluation dimensions in the prompt can improve performance on the epistemological dimensions. 
    \item We notice trade-offs between dimensions. Most notably, there seems to be a trade-off between epistemological and presentational quality.
    \item We analyze the relation of our evaluation and attribution-based evaluations of LLMs~\citep{rashkin2022measuring} and find that they emerge as mostly orthogonal and complementary.
\end{itemize}

\section{Evaluative Dimensions} %
\label{sec:dimensions}
Scholarship on science communication -- originating from disciplines such as communication science, sociology, psychology, human geography, and education, among others \citep{Trench.2021,Nisbet.2018,oxford-sciencecomm} -- offers conceptual arguments and empirical evidence for appropriately disseminating scientific information, e.g., on climate change, to the general public \citep{koenig_2023,Lewis_Neil_2021}. Building on this knowledge, we distinguish between two basic dimensions. (1) \emph{Presentational} features of the message that address its comprehensibility \citep{Lang.2000}. (2) \emph{Epistemological} features aiming to capture the degree to which the conveyed information represents current scientific knowledge adequately and comprehensively, while being specific and appropriately communicating associated uncertainties \citep{Fahnrich.2023}.

\subsection{Presentational Adequacy}
\label{sect:presentational-adequacy}
An adequate \emph{presentation} should comply with three criteria~\citep{oxford-sciencecomm}: (1) be comprehensible, (2) aid understanding through layout and visualizations, and (3) use appropriate sources and references. Here we focus primarily on \emph{comprehensibility}, evaluated along four criteria: style, clarity, linguistic correctness, and tone.

\textbf{Style.} The language should not be too informal or colloquial~\citep{Mazer.2008}, as this can undermine the credibility of information~\citep{Scharrer.2012}. Answers should not be too short or too long: brief snippets of information can lead to a ``feeling of knowing''~\citep{Leonhard.2020}, while long texts require motivation and cognitive resources that readers may not want to invest~\citep{Lang.2000}. In addition, we borrow some stylistic dimensions from the Multidimensional Quality Metrics (MQM) framework for  the evaluation of translations~\citep{mqm-lommel-2013}. 

\textbf{Clarity.} Responses should be concise and clearly formulated \citep{Maibach.2023}.  
The use of jargon and difficult technical content should be avoided~\citep{BaramTsabari.2013,BaramTsabari.2020}.

\textbf{(Linguistic) Correctness.} As in MQM, messages should adhere to linguistic conventions, i.e., the correct use of punctuation, spelling, and grammar.\footnote{ \url{https://themqm.info/typology}.} Violations can damage perceived credibility~\citep{Berger.2020,Mollick.2014}

\textbf{Tone.} The tone of a message concerns its perceived neutrality, its persuasiveness and its positivity or negativity. Science communication, especially climate-related, can be more effective if it doesn't lean towards a certain valence, worldview, or ideological conviction~\citep{Blanton.2019,yuan2020s,Kerr.2022,MunozCarrier.2020}. Likewise, messages should not use too positively or negatively valenced language, particularly if the goal is to convey factual information~\citep{palm2020don}.

\subsection{Epistemological Adequacy}
\label{sect:epistemic-adequacy}
The epistemological adequacy of climate-related messages is of greatest importance. This entails several aspects: (1) accuracy, (2) specificity, (3) completeness, (4) the degree of (un)certainty, and (5) the presentation of methods and methodology. Here we focus on the first four.

\textbf{Accuracy.} Scientific information should be \emph{accurate}~\citep{Kelesidou.2021}. This is crucial, considering known issues of LLMs such as \emph{hallucination}~\citep{Schafer.2023, Ji.2023}. We identify issues that deal with \emph{incorrect, wrong,} or \emph{self-contradictory} information, as well messages that take scientific findings, or anecdotal evidence, out of context~\citep{Hinnant.2016}. 

\textbf{Specificity.} Information that is \emph{relevant} to the audience should not be missed, while ignoring irrelevant information. Responses should address the \emph{spatial} and \emph{temporal} context; as specific, local information leads to higher perceived relevance~\citep{lee2015predictors,Leiserowitz.2017,Holmes.2020b}. In the absence of a specific time frame, the answer should generally be based on up-to-date knowledge.

\textbf{Completeness.} Rather than only referring to a part of the question posed, answers should be formulated in a way that addresses all aspects of the question in full~\citep{Leiserowitz.2017,bergquist2022information}. At the same time, the information given should reflect the depth and breadth of relevant scientific knowledge available regarding the topic(s) addressed~\citep{Kelesidou.2021}.

\textbf{Uncertainty.} Communicating the level of agreement and confidence regarding scientific findings, and supporting evidence, can be crucial to adequately informing the audience~\citep{howe2019acknowledging,budescu2012effective,Keohane.2014}. This is particularly important in climate communication~\citep{Maertens.2020,chinn2021effects,goldberg2022communicating}, scientific consensus on climate change has been found to function as a ``gateway belief'' and motivate public action~\citep{vanderLinden.2015}. 

\subsection{Aggregation of scores across dimensions}
In this paper we don't address the important question of how individual dimensions should be combined in a single metric, e.g., for model selection and benchmarking. This is a complex topic which requires assigning a value to each individual dimension. We also believe that the combination of these scores will vary by application.

\section{Human Evaluation Study}
\label{sec:experiments}
We test our evaluative dimensions in a human rating study. The rating task involves evaluating an answer based on the presentational  (\cref{sect:presentational-adequacy}) and epistemological dimensions (\cref{sect:epistemic-adequacy}). Screenshots of the template can be found in \cref{sect:appendix-template-screenshots}.
We select candidate raters with relevant educational background (see \cref{sect:appendix-raters-demographics}).
To be admitted, after finishing a brief tutorial, the raters need to pass an admission test (see \cref{sect:appendix-tutorial-and-admission}).
A summary of the broad demographics of participants can be found in \cref{sect:appendix-raters-demographics}.
Each answer is assessed by three human raters. We don't discourage brief consultations of external sources to clarify specific points but advise against extensive research.

\subsection{Question and Answer Data}
\label{sect:data}

\subsubsection{Questions}

A comprehensive evaluation would ideally cover a broad spectrum of information needs, including the basics of climate science, mitigation and adaptation, as well as context-specific issues; e.g., to address the concerns of vulnerable or under-resourced communities~\citep{amini2023ai}. 
However, no standardized tests exist to assess climate-related knowledge; in contrast to e.g. the medical domain~\citep{singhal2023large}. Hence, we begin by creating a diverse set of $300$ questions about topics that are either popular among search users, controversial or context-specific.

We collect questions from three different sources. 
For the first set, we use Google Trends, which provides data on public interest in specific search topics.\footnote{\url{https://trends.google.com/trends/}} We collect the most popular questions, by search volume, from the U.S., for the topics `Climate Change' and `Global Warming' for 2020-2022.
For the second set, we turn to Skeptical Science, a website that publishes authoritative information about climate science. We take the list of debated \emph{myths}\footnote{\url{https://skepticalscience.com/argument.php}} and manually rephrase them as questions.
Lastly, we use \gptfour{} to generate synthetic questions from the English Wikipedia.
We manually select a list of articles related to climate change (e.g., "Global Dimming", "Polar Amplification"), or discuss the topic in specific locations (e.g., "Climate Change in [COUNTRY]"), for a total of 139 articles. Then we split the documents in paragraphs and ask \gptfour{} to generate questions that can be answered by the paragraph. We apply several filters to assure that the Wikipedia questions are not overly dependent on the context and are therefore answerable only from the given paragraph. See \cref{sec:synthetic-questions} for more details and a discussion of filtering choices.

We post-process all questions to remove duplicates, questions that are not related to climate change, or taken out of context. Finally, we sample $100$ questions from each set. 

\subsubsection{Answers}

Generated answers can display a great deal of variation depending on prompt engineering, reasoning schemes, in-context learning, etc. However, a direct question is the most common way for users to get answers from LLMs. As such, a plain question provides a valuable baseline, reducing variance due to individual LLM's skills and optimization effort, and limiting confounding factors. To obtain answers we use a simple prompt consisting of the instruction: \textit{You are an expert on climate change communication. Answer each question in a 3-4 sentence paragraph.} We include the answer length information to anchor the expected response to an objective value. 

\subsection{Auxiliary Data}
\label{subsec:aux_data}
We support raters with AI Assistance, consisting of a model-generated critique for each evaluated dimension. For epistemological dimensions the assistance is grounded in verbatim evidence from relevant passages extracted from Wikipedia articles. To produce all necessary auxiliary data we carefully design a simple, robust baseline system (\cref{fig:sys-design}), which relies on a single LLM.  For consistency, we always use \gptfour{} for this purpose. Besides testing our evaluation we also run a comparison with an attribution-based evaluation (AIS)~\citep{rashkin2022measuring}, on the same data.

\textbf{Keypoints.} To find supporting evidence for an answer, for AI Assistance and AIS evaluation (\cref{sect:ais}), we extract keypoints from each answer. To do so, we instruct \gptfour{} to examine all the statements in the answer, and identify one to three key statements that are made in answering the question. We find this to provide better signal to retrieve evidence (see the next paragraph) than either using the whole answer or all sentences individually~\citep{liu2023evaluating}.

\textbf{Evidence.} For each keypoint we fetch evidence from Wikipedia. Given the question and the answer, we first ask \gptfour{} to provide the URL of a Wikipedia article that supports the answer. See \cref{tab:prompts} for the exact prompt. We limit evidence to Wikipedia because \gptfour{} is fairly consistent in generating relevant, valid Wikipedia URLs, while the quality is lower for the unrestricted web. Furthermore, Wikipedia is uniform in style and quality as it adheres to established guidelines.\footnote{\url{https://en.wikipedia.org/wiki/Wikipedia:Policies_and_guidelines}.} While random web pages can vary significantly in content and presentation quality.

We break down the relevant article into its paragraphs. For each keypoint, we ask the model to score the paragraphs based on their relevance to the keypoint and the question. We pick the $3$ highest scoring ones as evidence (cf. \cref{tab:ais-example} for an example). We find that using keypoints, in combination with URL generation and evidence selection, is a simple and effective solution. In particular, we find this to work better than off-the-shelf sparse or dense retrieval (e.g., using BM25/GTR~\citep{gtr-ni-etal-2022-large}) over Wikipedia passages.

\textbf{AI Assistance.} To assist human raters, we use \gptfour{} to critique the answer along the dimensions introduced in \cref{sec:dimensions}. For each dimension, we ask the model to express its agreement or disagreement that the information is presented well according to that dimension. For epistemological dimensions, we also provide the retrieved evidence and instruct the model to quote the evidence verbatim to support its disagreement (if any).

Please refer to \cref{tab:prompts} for a complete list of prompts used to generate the data, and to \cref{sec:answer-stats} for some statistics of the generated answers.

\begin{figure*}[t]
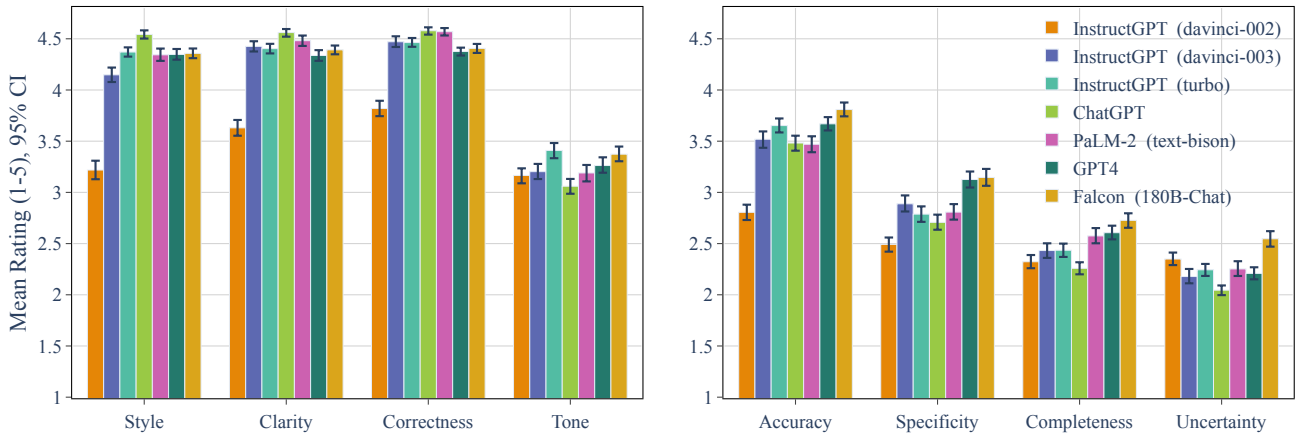

     \centering
     \begin{subfigure}[T]{0.495\textwidth}
         \centering
         \includesvg[width=\textwidth]{figures/main-presentational.svg} %
     \end{subfigure}
     \hfill
     \begin{subfigure}[T]{0.495\textwidth}
         \centering
         \includesvg[width=\textwidth]{figures/main-epistemic.svg} %
     \end{subfigure}
    \caption{Bootstrapped mean rating, and 95\% confidence intervals, for all presentational (left) and epistemological (right) dimensions.} \label{fig:all_results}
\end{figure*}

\section{Experimental Results}
Here we present the findings from the experiments using \cref{fig:all_results} as a summary. Full results tables, including confidence intervals, are reported in \cref{tab:main_results_presentational} and \cref{tab:main_results_epistemological}. We also report pairwise LLM t-tests in \cref{tab:pairwise_presentational,tab:pairwise_epistemological}.  We compute pairwise distance and Krippendorff's alpha agreement metrics for all experiments in \cref{sec:agreement}, including an analysis of rating timing \cref{sec:timing}. Accurate rating of climate information is challenging, but we find the main conclusion proposed below to be adequately supported.

\textbf{LLMs.} We evaluate the following models: \gptfour{}~\citep{openai2023gpt4}, \gptthreex{}, \gptturbo{}, \gptthree{}, \gpttwo{}\footnote{\url{https://platform.openai.com/docs/models}.}, as well as \palm{}~\citep{anil2023palm} and ~\falcon{}\footnote{\url{https://falconllm.tii.ae/falcon.html}.}. This data was collected in the months of September and October 2023.

\subsection{Performance Results}
\textbf{Presentational Results.} 
Overall, except for \gpttwo{}, LLMs produce clear, fluent, linguistically correct text. 
This confirms how far LLMs have come in terms of surface form quality, seemingly thanks to RLHF~\citep{instruction-tuning-ouyang2022training}. We note, however, a marked performance drop for \emph{tone}. This suggests that the evaluation of LLM's presentation should probably shift its focus on subtler aspects of language use (cf. also \cref{subsect:appendix-emerging-patterns}).

\textbf{Epistemological Results.} 
Compared to presentation, the epistemological evaluation reveals lower performance across all models and dimensions. Results are consistently low for the last three: \textit{specificity}, \textit{completeness} and \textit{uncertainty}. 
We note that these dimensions may be difficult to satisfy in short 3-4 sentence answers. Being comprehensive in such a short space may be harder than being accurate.

On the other hand, LLMs don't seem to make a good use of space (see \cref{subsect:appendix-emerging-patterns}). Thus, space constraints alone do not seem sufficient to explain the result.
Overall, on climate information, current top-of-the-line LLMs have significant headroom for improvement. For examples, please see Tables~\ref{tab:example_accuracy} to \ref{tab:example_uncertainty}.

\textbf{Dimension-Aware Prompts.}
In a follow-up experiment, using only \gptfour{}\footnote{This experiment was carried out in November 2023, after a major release from OpenAI, on Nov 6. \gptfour{}'s performance cannot be directly compared with the previous results, because \gptfour{} is also used to produce the auxiliary data.}, we found that including a description of the evaluation criteria in the prompt can improve performance on the difficult dimensions: epistemological and \textit{tone}. \cref{tab:eval-aware-prompt} compares \gptfour{} with either the 'basic' or 'dimension-aware' prompts (see \cref{tab:prompts} for the actual prompt's text). Interestingly, better performance on "knowledge" comes at the cost of worse quality in the presentational dimensions (except for \emph{tone}), providing additional evidence for the existence of intrinsic tradeoffs between surface and epistemological quality.
\begin{table}
\begin{center}
\begin{tabular}{lcc}\toprule
        Dimension     & Basic prompt & Augmented prompt \\\midrule
         Style        & 4.33 & 4.10\\
         Clarity      & 4.34 & 4.24\\
         Correctness  & 4.43 & 4.29\\
         Tone         & 3.34 & 3.59\\\midrule
         Accuracy     & 3.77 & 3.92\\
         Specificity  & 3.29 & 3.46\\
         Completeness & 2.72 & 2.97\\
         Uncertainty  & 2.73 & 2.92\\\bottomrule
    \end{tabular}
\caption{\gptfour{} with basic vs. evaluation-aware prompts.}
\label{tab:eval-aware-prompt}
\end{center}
\end{table}

\subsection{Resolution and Range.}
The results suggest that the evaluation possesses sufficient resolution to distinguish models on specific dimensions, indicate where they differ and suggest trends. For instance, \chatgpt{} is the best on all presentation dimensions, except \emph{tone}, but ranks between position 4 and last on the epistemological scores. This indicates the possibility of complex dependencies between presentational and epistemological properties.
As a matter of fact, \gptfour{} is always better than \chatgpt{} across the epistemological dimensions but worse on most presentational ones.

The difference between the best LLM and the worst in specific dimensions -- e.g., \falcon{} and \gpttwo{} on the epistemological ones, and, respectively, \chatgpt{} and \gpttwo{} on the first three presentational ones--is large compared to the estimated standard deviation, providing evidence that the evaluation has sufficient dynamic range.
Noticeably, the most competitive model--at the time of writing--on the epistemological dimensions is an open model, \falcon{}. This LLM's performance may be related to its large size, but we can only speculate as this information is not generally available for all models. 

\subsection{Influence of Question Source.}
Comparing the rating outcome by source of the question – Skeptical Science, GTrends, and synthetic questions from Wikipedia-–we find no major differences, except for slightly lower scores for Wikipedia questions. In particular, we see no evidence that \gptfour{} performs better on questions that were generated with \gptfour{} compared to the other sources. Similarly, the topic of the question does not show a strong correlation with answer quality. See \cref{sec:rating_by_type} for additional discussion and figures.

\subsection{Pragmatic Style}
\label{subsect:appendix-emerging-patterns}
\begin{table}
\footnotesize
\begin{tabular}{@{}p{8cm}}
\toprule
{\bf Question}: Are we heading towards an ice age?\\\midrule
{\bf Answer}: No, we are not heading towards an ice age. \ctext{clipyellow}{The current scientific consensus is that human activities, particularly the emission of greenhouse gases, are causing the planet to warm at an unprecedented rate. This rapid warming is leading to a variety of climate changes, including more frequent and severe heatwaves, rising sea levels, and changes in precipitation patterns.} While natural processes have triggered ice ages in the past, these are very slow processes that take thousands of years, and the current rapid warming trend is not consistent with the onset of a new ice age.\\\bottomrule
\end{tabular}
\vspace{-8pt}
\caption{Example of common LLMs' argumentative style.}
\label{tab:example-style}
\end{table}

LLMs often fill parts of the answer with somewhat relevant but unspecific, cautionary and sometimes preachy reminders. The example in \cref{tab:example-style} illustrates the point.
Due to the highlighted part, the response may result in sub-optimal \emph{cooperative conversation}, in a Gricean's sense~\citep{levinson_1983}. For instance, one could argue that the \emph{maxim of quantity} is being violated ('do not provide more information than required') as most of what follows the first sentence is strictly speaking unnecessary.
The \emph{maxim of manner} ('be relevant') may also be violated: comments on extreme weather and rising sea levels are only loosely related to the question. That space could be used to provide more specific information.

Furthermore, the answer relies generically on the notion of \emph{scientific consensus}, which happens relatively frequently in our data. Besides the possibility of being superficially interpreted as an `argument from authority', research suggests \citep{orchinik23} that the `consensus' argument can be surprisingly ineffective due to complex belief system underlying how such arguments are processed.  
\citet{orchinik23} argue that perceived scientists credibility, which in turn may depend on general worldview, affects how consensus-based messages are received and receptiveness to future messaging. 
This presentation style may not appeal to the different audiences of science communication and possibly lead to diminished interest and fatigue~\citep{schaefer-et-al-2018}. To further complicate the picture, recent work points out that after conversing with AI on climate change, people with a skeptical stance end up dissatisfied but also more supportive of scientific consensus~\citep{chen24}

In these respects, LLMs answers differ from some human experts' answers to similar questions.\footnote{E.g., from \url{https://climatefeedback.org/} or \url{https://skepticalscience.com/}.} The latter tend to rely on direct and specific scientific evidence; e.g., in the case of the question above, an expert may cite land, atmospheric and ocean data for temperature trends, from multiple scientific sources.\footnote{Human experts' answers also tend to include images summarizing quantitative data.}
Our framework captures some of these aspects in dimensions like \emph{tone} and \emph{specificity}, but the pragmatics aspects of Generative AI should probably be investigated more directly in the future.

\subsection{Role of AI Assistance.}
We expect human raters to identify more (real) issues with assistance,
because it makes them aware of them. We find supporting evidence in two separate experiments.
\begin{figure}%
    \centering
    \includesvg[width=\columnwidth]{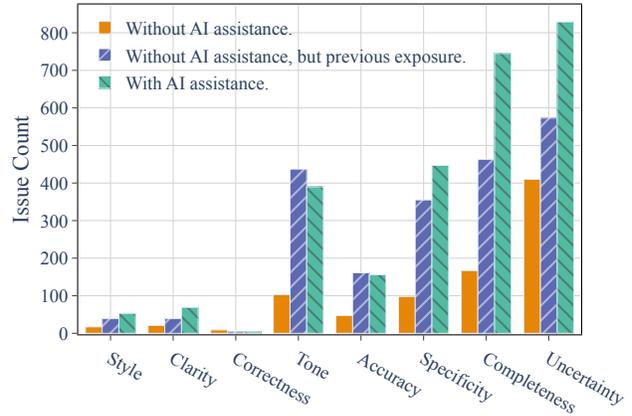}
    \caption{Number of issues detected depending on\\AI Assistance exposure.}
    \label{fig:critique-exposure-a}
\end{figure}

\cref{fig:critique-exposure-a} reports the number of issues detected for each dimension on \gptfour{} answers in three different settings, each with a different degree of the raters' exposure to assistance. 'Without AI Assistance' refers to a setting where a specific pool of raters is never exposed to rating with AI Assistance. 
'Without AI Assistance, but previous exposure' refers to a setting where no assistance was shown, but the raters have worked on previous studies that included assistance.\footnote{We do make sure that the raters have not worked on the same examples before and have never seen assistance for the specific examples they are working on.}
Lastly, 'With AI Assistance' denotes the standard setting where assistance is shown anytime is available.

Results suggest that the presence of assistance is key for detecting more issues. 
This is consistent with the results from \citet{saunders2022selfcritiquing}.
Raters with previous exposure to assistance are in a ``middle'' position: They detect more issues than the assistance-unaware group, but less than the group provided with assistance for the experiment. This suggests that raters learn from repeated exposure to assistance, and show improved performance even when no assistance is present.

Further evidence of the usefulness of AI Assistance comes from our validation experiments (cf. \cref{sect:validation-details} for more details).  Similar to \citet{saunders2022selfcritiquing}, we want to determine if assistance helps surface real issues, without general access to gold truth in our data. To do this, the authors manually generated $30$ different examples, each exhibiting a particular issue. We found that the majority of three raters detected $77\%$ of issues when shown assistance, while the majority of three raters only detected $60\%$ of the issues when not shown assistance.

\begin{figure}[tbh]
    \centering
    \includesvg[width=0.99\columnwidth]{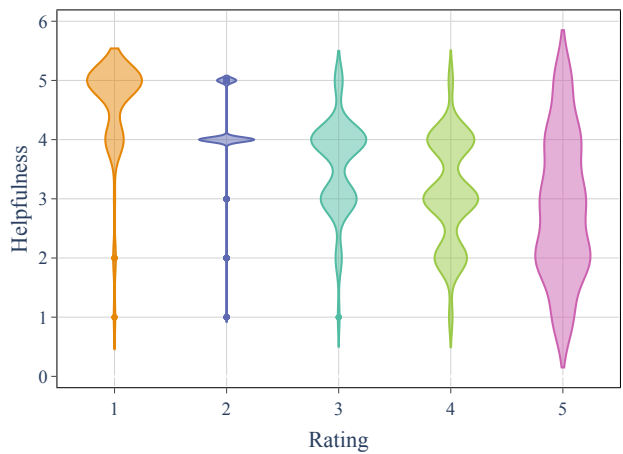}
    \setlength{\belowcaptionskip}{-20pt}
    \vspace{-5pt}
    \caption{The relationship between rating and reported helpfulness of the AI assistance (on the same scale).}
    \label{fig:critique-exposure-b}
\end{figure}

In our experiments we collected feedback from raters on the helpfulness of assistance. The data suggests that when raters do not find assistance helpful, they give higher ratings (see \cref{fig:critique-exposure-b}). This indicates that the raters can think critically about the assistance and do not follow it blindly. %
These experiments provide evidence that the AI Assistance helps the raters find real issues that they would not have otherwise been reported.

\subsection{Epistemological Adequacy and Attribution}
\label{sect:ais}
Grounding LLMs responses in retrieved documents, or Retrieval Augmented Language Models (RALM) \citep{pmlr-v119-guu20a,NEURIPS2020_6b493230}, has been proposed to improve LLMs' response quality and alleviate factuality limitations~\citep{gophercite}. Analogously, on the evaluation side,
frameworks such as Attribution to Identified Source (AIS) argue in favour of dedicated evaluations that bypass difficult direct factuality assessements~\citep{rashkin23, dziri2022evaluating}: an attributable answer must include an explicit quote, from an existing document. 
AIS signals can be also modeled automatically \citep{bohnet2023attributed} enabling training via reinforcement learning~\citep{roit-etal-2023-factually}. 

While evaluating the ability of LLMs to properly ground their statements goes beyond the scope of this paper\footnote{For instance, as proposed by \citet{liu2023evaluating}, this may involve evaluating generative search engines.}, we begin examining the relationship between attribution and the epistemological dimensions with an AIS experiment. We run this experiment only on \gptfour{}. 

In our data, each answer is associated with a set of keypoints which, in turn, are used to identify Wikipedia articles that are likely to contain supporting evidence. For 87.7\%  of the questions, \gptfour{} produces a valid Wikipedia article from which evidence passages can be extracted. 
We evaluate the attribution of each keypoint individually by asking the human annotators whether a keypoint is fully, partially or not supported by the evidence. $66.79\%$ of keypoints are either fully or partially supported. We consider an answer to be fully attributed if all its keypoints are supported. An answer is not supported if all its keypoints are not supported
At the answer level, $46.08\%$ of the answers are fully or partially supported by the evidence. While providing only preliminary evidence, the data suffices for a first analysis.

\begin{figure}
     \centering
         \includesvg[width=1.0\columnwidth]{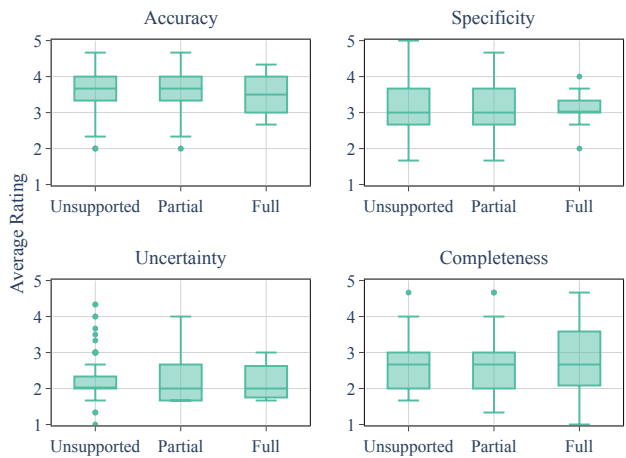}
\caption{Comparing AIS ratings with average ratings of the $4$ epistemological dimensions.
        }
\vspace{-5pt}
\label{fig:ais}
\end{figure}

\cref{fig:ais} compares the distribution of average epistemological ratings, with respect to the attribution of answers, revealing interesting trends. In both the \emph{accuracy} and \emph{specificity} dimensions, we observe that answers that are fully attributed have higher minimum ratings compared to answers that are only partially attributed, or not attributed at all. Interestingly, we see an opposite pattern in the \emph{completeness} dimension: Answers that are fully attributed have lower minimum ratings on \emph{completeness}. This result highlights a blind spot for attribution methods; AIS can only consider what \emph{is} included in the answers, and not what important information is missing. In the \emph{uncertainty} dimension, we observe that there are more answers with low uncertainty ratings among the answers that are not attributed, compared to answers that are either partially or fully attributed. 

More generally, there does not seem to be any correlation between AIS and epistemological results. The Spearman's coefficient between AIS and the 3-raters mean rating value for \emph{accuracy}, \emph{specificity}, \emph{uncertainty} and \emph{completeness} are, respectively: $0.03$, $-0.06$, $0.002$, $-0.02$, with corresponding p-values: $0.65, 0.31, 0.97, 0.78$.
We interpret this as evidence that AIS and epistemological assessments are mostly orthogonal.  We provide a few examples in \cref{tab:ais-examples}, in particular, of answers that are fully attributable but score low on epistemological dimensions. This suggests that, while practical and complementary, attribution, either human or model-based, is not a substitute for direct epistemological assessment. 

Science and climate communication is more likely to be trusted if the source is perceived as credible, engaged and concerned about the audience's interests~\citep{Brown.2011,Maibach.2023,hayhoe2018}. An adequate presentation of climate information should include curated references. In future work we plan to extend our framework to evaluate references in a principled, systematic ways.

\section{Limitations and Future Work}
\label{sect:limitations-future-work}
While our agreement analysis (\cref{sec:agreement}) suggests that the evaluation is robust at the system level, the rating dimensions inherently have a subjective component, introducing noise when evaluating at the answer-level. As we do not have access to gold ratings, calibration of raters remain an open issue, as reflected by the medium inter-rater agreement discussed in \cref{sec:agreement}. Future work should consider explicitly addressing this subjectivity in the data collection process (cf. \citet{rottger22}).

AI Assistance is an essential part of our evaluation, because it helps raters identify issues in the answers, particularly for the epistemological dimensions. As \cref{tab:main_results_presentational,tab:main_results_epistemological} shows, raters would fail to recognize many issues without the AI Assistance ('\gptfour{} no assistance'). However, the assistance may also influence the raters beyond enhancing discovery. It may only help in the discovery of some issues but not others. There may also be errors caused by models falsely pointing out issues and wrongly convincing the raters. The issues identified will likely vary by model. There is definitely a need to better understand these issues and to identify mitigation strategies. This links this research to the broader AI alignment field and will be one of the main focuses of our future work.
A related topic is the role of LLMs as raters. Preliminary experiments are promising (\cref{sect:appendix-llm-rater}). We found that, as with humans, LLMs benefit from AI Assistance and that humans and LLM raters tend to agree on major points.

Ideally, an answer would be tailored towards the audience, and take into account their specific attributes ~\citep{Hendriks.2016,Klinger.2021}. Unless specifically prompted, LLMs do not do this and the evaluation of such setting would introduce additional challenges.
Another important area for future work concerns multimodal responses.
Research provides abundant evidence on the importance of supplementing textual information with visual aids. %
\citep{flemming2018emotionalization, Brown.2011}. Visual complements can be especially useful for understanding quantitative data~\citep{Fagerlin.2011} and in the case of limited literacy~\citep{Wolf.2010}. The abstract nature of climate change, and its distant implications, makes visualization particularly challenging~\citep{Schafer.2020}.

\section{Related Work}
\textbf{Evaluating LLMs.} While LLMs can generate fluent text, responses are not always adequately grounded, attributable to reliable sources, and complete. For instance, \citet{liu2023evaluating} assess four generative search engines and report that, although responses are perceived as high quality, only half are fully supported. Their findings reveal an inverse correlation between fluency/utility and evidential support. \citet{xu2023critical} advocate for expert-level human evaluations in question answering, cautioning against over-reliance on single metrics instead of comprehensive assessments.

Another domain that needs expert-level evaluation is the medical domain. \citet{singhal2023large} propose Med-PaLM, an LLM for medical information, and introduces a clinical evaluation framework which covers criteria like alignment with scientific consensus, potential harm, and comprehension. Evaluating LLMs on climate information is another domain that can benefit from expert-level evaluation. However, prior work mainly focused on text classification tasks~\citep{climatefever, climatext, cards,pirav1,climatebert,cheaptalk,climabench,climatex}. This study aims to fill this gap by providing a comprehensive evaluation framework for generative climate information.

\textbf{Scalable Oversight.} This area, introduced by \citet{Amodei16}, studies the question of how to scale human oversight, especially in the setting where evaluating (or supervising) models becomes increasingly difficult.
Contributions have initially focused on theoretical proposals for how AI can help humans supervise models that exceed their abilities \citep{Irving18, Leike18, Christiano18}. 
Following \citet{Irving18}, one can see our AI Assistance as a single-turn debate, where the human annotator is shown the answer proposed by the model and a single response to that answer.\footnote{In the setting of \citet{Irving18}, this corresponds to the second level of the polynomial hierarchy $\Sigma_2^P$.}

Two recent studies provide interesting proofs of concepts for AI Assistance:
\citet{bowman2022measuring} study \emph{sandwiching}, an approach where non-experts align a model with the help of a model while experts provide validation. They show that non-expert raters perform better on an (artificially) difficult multiple-choice task when interacting with a dialogue agent.
Several studies also evaluated short debates in this setting with mixed results \citep{parrish23a, parrish23b, michael23}.
\citet{saunders2022selfcritiquing} report that human raters of summarization tasks produce more critiques when given the opportunity to accept or edit critiques written by a model. Our work contributes a study of a \emph{scalable oversight} protocol to improve rating quality in a realistic setting.

\textbf{AI Ratings.} Recent studies explore the feasibility of evaluations performed by AI.
\citet{gemba} indicate that LLMs can perform state-of-the-art quality assessment of translations, even without references. Their work has been extended to automatic MQM annotation by \citet{fernandes2023devil}. \citet{gilardi} reports that ChatGPT has a higher agreement with expert-level raters than with less qualified ones. \citet{chiang2023can} argue that humans and LLMs ratings are correlated but point out LLM's factuality and bias limitations. Instead of replacing human raters entirely, in our work we demonstrate the effectiveness of using AI Assistance to aid educated raters.

\section{Conclusion}
We introduce an evaluation framework informed by science communication research and assess LLMs on a first set of climate information needs. The task is difficult for human raters. To support them, an important part of our framework relies on a novel and practical protocol for scalable oversight that leverages AI Assistance. It is important to realize that these are the first results of this kind and more research is needed. In particular, while there is evidence that AI Assistance is valuable, we need to develop a framework to understand and mitigate undesired influence on the raters. Overall, our results suggest that, while presentationally adequate, current LLMs have much room for improvement regarding the epistemological qualities of their outputs. More research is needed to understand and improve these aspects of LLMs. %

\section*{Impact Statement}

In this work we present an evaluation framework to assess the quality of answers to climate-related questions. Our evaluation is based on science-communication principles and aims to evaluate responses to genuine information needs of the public. Progress in correctly answering such questions can have a large impact for the dissemination of scientific results and can lead to positive effects on climate literacy, also reducing the public's susceptibility to misinformation.

As with any evaluation there are however limits to its validity. Specifically, the evaluation of systems to be deployed in critical contexts requires additional grounding and expert verification. This is especially the case when system responses inform actions. Moreover, the evaluation is limited to the evaluated context, and we make no claims that models can be trusted and deployed outside of that context.

\section*{Acknowledgements}
We would like to thank the following people for their feedback and support: Leslie Leuenberger, Claire Foulquier-Gazagnes, Yana Koroleva, Srini Narayanan, Andrew Novikov, Annalisa Pawlosky, Fernando Pereira, Rachana Jayaram, Maria Ryabtseva.
We also extend our appreciation to the anonymous reviewers for their valuable suggestions and feedback.
\newpage
\bibliography{bib/sample}
\bibliographystyle{bib/icml2024}

\newpage
\appendix
\onecolumn
\section{Appendix}
\subsection{Main Results}
\cref{tab:main_results_presentational,tab:main_results_epistemological} report the full results for all evaluated models.

\begin{table}[htbp]
\centering
\small
\begin{tabular}{ll@{\hskip 3pt}ll@{\hskip 3pt}ll@{\hskip 3pt}ll@{\hskip 3pt}l}
\toprule
System & \multicolumn{2}{c}{style} & \multicolumn{2}{c}{clarity} & \multicolumn{2}{c}{correctness} & \multicolumn{2}{c}{tone}\\
\midrule
ChatGPT & 4.54 &\scriptsize [4.50, 4.58]& 4.56 &\scriptsize [4.52, 4.60]& 4.58 &\scriptsize [4.54, 4.61]& 3.06 &\scriptsize [2.99, 3.13] \\
InstructGPT (davinci-003) & 4.15 &\scriptsize [4.08, 4.22]& 4.43 &\scriptsize [4.38, 4.47]& 4.47 &\scriptsize [4.42, 4.52]& 3.20 &\scriptsize [3.12, 3.28] \\
InstructGPT (davinci-002) & 3.22 &\scriptsize [3.13, 3.31]& 3.63 &\scriptsize [3.55, 3.70]& 3.82 &\scriptsize [3.74, 3.90]& 3.17 &\scriptsize [3.09, 3.24]\\
InstructGPT (turbo) & 4.37 &\scriptsize [4.32, 4.42]& 4.40 &\scriptsize [4.36, 4.45]& 4.46 &\scriptsize [4.42, 4.51]& 3.41 &\scriptsize [3.33, 3.48] \\
PaLM-2 (text-bison) & 4.34 &\scriptsize [4.28, 4.40]& 4.48 &\scriptsize [4.43, 4.53]& 4.57 &\scriptsize [4.53, 4.61]& 3.19 &\scriptsize [3.11, 3.27] \\
GPT4 & 4.35 &\scriptsize [4.30, 4.40]& 4.34 &\scriptsize [4.28, 4.39]& 4.38 &\scriptsize [4.34, 4.41]& 3.26 &\scriptsize [3.19, 3.34] \\
Falcon (180B-Chat) & 4.36 &\scriptsize [4.31, 4.41]& 4.39 &\scriptsize [4.35, 4.44]& 4.41 &\scriptsize [4.36, 4.45]& 3.37 &\scriptsize [3.30, 3.45] \\
\midrule
GPT4, no assistance, prev.\ exposure & 4.59 &\scriptsize [4.54, 4.63]& 4.63 &\scriptsize [4.59, 4.68]& 4.66 &\scriptsize [4.63, 4.70]& 3.24 &\scriptsize [3.16, 3.32] \\
GPT4, no assistance & 4.45 &\scriptsize [4.41, 4.50]& 4.57 &\scriptsize [4.53, 4.61]& 4.74 &\scriptsize [4.70, 4.77]& 4.35 &\scriptsize [4.29, 4.42] \\
\bottomrule
\end{tabular}
    \caption{Results along the presentational dimensions, with $95\%$ confidence intervals.}
    \label{tab:main_results_presentational}
\end{table}

\begin{table}[htbp]
\centering
\small
\begin{tabular}{ll@{\hskip 3pt}ll@{\hskip 3pt}ll@{\hskip 3pt}ll@{\hskip 3pt}l}
\toprule
System & \multicolumn{2}{c}{accuracy} & \multicolumn{2}{c}{specificity} & \multicolumn{2}{c}{completeness} & \multicolumn{2}{c}{uncertainty}\\
\midrule
ChatGPT & 3.48 &\scriptsize [3.41, 3.55]& 2.71 &\scriptsize [2.63, 2.78]& 2.26 &\scriptsize [2.20, 2.31]& 2.05 &\scriptsize [2.00, 2.09] \\
InstructGPT (davinci-003) & 3.52 &\scriptsize [3.44, 3.60]& 2.89 &\scriptsize [2.81, 2.97]& 2.43 &\scriptsize [2.36, 2.50]& 2.18 &\scriptsize [2.11, 2.25] \\
InstructGPT (davinci-002) & 2.81 &\scriptsize [2.73, 2.88]& 2.49 &\scriptsize [2.42, 2.56]& 2.32 &\scriptsize [2.26, 2.39]& 2.35 &\scriptsize [2.29, 2.41] \\
InstructGPT (turbo) & 3.65 &\scriptsize [3.58, 3.73]& 2.79 &\scriptsize [2.71, 2.86]& 2.43 &\scriptsize [2.37, 2.50]& 2.24 &\scriptsize [2.19, 2.30] \\
PaLM-2 (text-bison) & 3.47 &\scriptsize [3.39, 3.55]& 2.81 &\scriptsize [2.73, 2.89]& 2.57 &\scriptsize [2.50, 2.65]& 2.25 &\scriptsize [2.18, 2.32] \\
GPT4 & 3.67 &\scriptsize [3.61, 3.73]& 3.13 &\scriptsize [3.05, 3.21]& 2.61 &\scriptsize [2.53, 2.68]& 2.21 &\scriptsize [2.15, 2.27] \\
Falcon (180B-Chat) & 3.81 &\scriptsize [3.74, 3.87]& 3.15 &\scriptsize [3.07, 3.23]& 2.73 &\scriptsize [2.65, 2.80]& 2.55 &\scriptsize [2.47, 2.62] \\
\midrule
GPT4, no assistance, prev.\ exposure & 3.86 &\scriptsize [3.79, 3.93]& 3.43 &\scriptsize [3.35, 3.52]& 3.30 &\scriptsize [3.21, 3.39]& 2.78 &\scriptsize [2.69, 2.87] \\
GPT4, no assistance & 4.49 &\scriptsize [4.44, 4.55]& 4.41 &\scriptsize [4.35, 4.48]& 4.32 &\scriptsize [4.25, 4.39]& 3.38 &\scriptsize [3.29, 3.46] \\
\bottomrule
\end{tabular}
    \caption{Results along the epistemological dimensions, with $95\%$ confidence intervals.}
    \label{tab:main_results_epistemological}
\end{table}

\subsection{Pairwise t-tests}
\cref{tab:pairwise_presentational,tab:pairwise_epistemological} report the results of LLM pairwise t-tests.

\begin{table}[htbp]
\centering
\small
\begin{tabular}{llccccccc}
\toprule
 && \multicolumn{3}{c}{InstructGPT}  & \multicolumn{1}{c}{ChatGPT} & \multicolumn{1}{c}{PaLM-2} & \multicolumn{1}{c}{GPT4} & \multicolumn{1}{c}{Falcon}  \\
& \multicolumn{1}{c}{\scriptsize davinci-002} & \multicolumn{1}{c}{\scriptsize davinci-003} & \multicolumn{1}{c}{\scriptsize turbo} & \multicolumn{1}{c}{} & \multicolumn{1}{c}{\scriptsize text-bison} & \multicolumn{1}{c}{} & \multicolumn{1}{c}{\scriptsize 180B-Chat}  \\

\midrule
\multicolumn{2}{l}{\bf style} \\
& InstructGPT (davinci-002) &  & $--$ & $--$ & $--$ & $--$ & $--$ & $--$ \\
& InstructGPT (davinci-003) & $++$ &  & $--$ & $--$ & $--$ & $--$ & $--$ \\
& InstructGPT (turbo) & $++$ & $++$ &  & $--$ & $\sim$ & $\sim$ & $\sim$ \\
& ChatGPT & $++$ & $++$ & $++$ &  & $\sim$ & $\sim$ & $\sim$ \\
& PaLM-2 (text-bison) & $++$ & $++$ & $\sim$ & $\sim$ &  & $\sim$ & $\sim$ \\
& GPT4 & $++$ & $++$ & $\sim$ & $\sim$ & $\sim$ &  & $\sim$ \\
& Falcon (180B-Chat) & $++$ & $++$ & $\sim$ & $\sim$ & $\sim$ & $\sim$ &  \\

\midrule
\multicolumn{2}{l}{\bf clarity} \\
& InstructGPT (davinci-002) &  & $--$ & $--$ & $--$ & $--$ & $--$ & $--$ \\
& InstructGPT (davinci-003) & $++$ &  & $\sim$ & $--$ & $\sim$ & $\sim$ & $\sim$ \\
& InstructGPT (turbo) & $++$ & $\sim$ &  & $--$ & $-$ & $\sim$ & $\sim$ \\
& ChatGPT & $++$ & $++$ & $++$ &  & $\sim$ & $\sim$ & $\sim$ \\
& PaLM-2 (text-bison) & $++$ & $\sim$ & $+$ & $\sim$ &  & $\sim$ & $\sim$ \\
& GPT4 & $++$ & $\sim$ & $\sim$ & $\sim$ & $\sim$ &  & $\sim$ \\
& Falcon (180B-Chat) & $++$ & $\sim$ & $\sim$ & $\sim$ & $\sim$ & $\sim$ &  \\

\midrule
\multicolumn{2}{l}{\bf correctness} \\
& InstructGPT (davinci-002) &  & $--$ & $--$ & $--$ & $--$ & $--$ & $--$ \\
& InstructGPT (davinci-003) & $++$ &  & $\sim$ & $--$ & $--$ & $\sim$ & $\sim$ \\
& InstructGPT (turbo) & $++$ & $\sim$ &  & $--$ & $--$ & $\sim$ & $\sim$ \\
& ChatGPT & $++$ & $++$ & $++$ &  & $\sim$ & $\sim$ & $\sim$ \\
& PaLM-2 (text-bison) & $++$ & $++$ & $++$ & $\sim$ &  & $\sim$ & $\sim$ \\
& GPT4 & $++$ & $\sim$ & $\sim$ & $\sim$ & $\sim$ &  & $\sim$ \\
& Falcon (180B-Chat) & $++$ & $\sim$ & $\sim$ & $\sim$ & $\sim$ & $\sim$ &  \\

\midrule
\multicolumn{2}{l}{\bf tone} \\
& InstructGPT (davinci-002) &  & $\sim$ & $--$ & $\sim$ & $\sim$ & $-$ & $--$ \\
& InstructGPT (davinci-003) & $\sim$ &  & $--$ & $\sim$ & $\sim$ & $\sim$ & $--$ \\
& InstructGPT (turbo) & $++$ & $++$ &  & $\sim$ & $\sim$ & $\sim$ & $\sim$ \\
& ChatGPT & $\sim$ & $\sim$ & $\sim$ &  & $-$ & $--$ & $--$ \\
& PaLM-2 (text-bison) & $\sim$ & $\sim$ & $\sim$ & $+$ &  & $\sim$ & $--$ \\
& GPT4 & $+$ & $\sim$ & $\sim$ & $++$ & $\sim$ &  & $-$ \\
& Falcon (180B-Chat) & $++$ & $++$ & $\sim$ & $++$ & $++$ & $+$ &  \\

\bottomrule
\end{tabular}
    \caption{Presentational dimensions - Pairwise t-test results. $--$ and $++$ indicate high significance ($p<0.01$) that the model in the row scores lower/higher than the model in the colum. Single $-/+$ indicate $p < 0.05$ and $\sim$ indicates no significant difference between the models.}
    \label{tab:pairwise_presentational}
\end{table}

\begin{table}[htbp]
\centering
\small
\begin{tabular}{llccccccc}
\toprule
 && \multicolumn{3}{c}{InstructGPT}  & \multicolumn{1}{c}{ChatGPT} & \multicolumn{1}{c}{PaLM-2} & \multicolumn{1}{c}{GPT4} & \multicolumn{1}{c}{Falcon}  \\
& \multicolumn{1}{c}{\scriptsize davinci-002} & \multicolumn{1}{c}{\scriptsize davinci-003} & \multicolumn{1}{c}{\scriptsize turbo} & \multicolumn{1}{c}{} & \multicolumn{1}{c}{\scriptsize text-bison} & \multicolumn{1}{c}{} & \multicolumn{1}{c}{\scriptsize 180B-Chat}  \\

\midrule
\multicolumn{2}{l}{accuracy} \\
& InstructGPT (davinci-002) &  & $--$ & $--$ & $--$ & $--$ & $--$ & $--$ \\

& InstructGPT (davinci-003) & $++$ &  & $--$ & $\sim$ & $\sim$ & $--$ & $--$ \\

& InstructGPT (turbo) & $++$ & $++$ &  & $\sim$ & $\sim$ & $\sim$ & $--$ \\

& ChatGPT & $++$ & $\sim$ & $\sim$ &  & $\sim$ & $--$ & $--$ \\

& PaLM-2 (text-bison) & $++$ & $\sim$ & $\sim$ & $\sim$ &  & $--$ & $--$ \\

& GPT4 & $++$ & $++$ & $\sim$ & $++$ & $++$ &  & $--$ \\

& Falcon (180B-Chat) & $++$ & $++$ & $++$ & $++$ & $++$ & $++$ &  \\

\midrule
\multicolumn{2}{l}{specificity} \\
& InstructGPT (davinci-002) &  & $--$ & $--$ & $--$ & $--$ & $--$ & $--$ \\

& InstructGPT (davinci-003) & $++$ &  & $\sim$ & $\sim$ & $\sim$ & $--$ & $--$ \\

& InstructGPT (turbo) & $++$ & $\sim$ &  & $\sim$ & $\sim$ & $--$ & $--$ \\

& ChatGPT & $++$ & $\sim$ & $\sim$ &  & $-$ & $--$ & $--$ \\

& PaLM-2 (text-bison) & $++$ & $\sim$ & $\sim$ & $+$ &  & $--$ & $--$ \\

& GPT4 & $++$ & $++$ & $++$ & $++$ & $++$ &  & $\sim$ \\

& Falcon (180B-Chat) & $++$ & $++$ & $++$ & $++$ & $++$ & $\sim$ &  \\

\midrule
\multicolumn{2}{l}{completeness} \\
& InstructGPT (davinci-002) &  & $-$ & $--$ & $\sim$ & $--$ & $--$ & $--$ \\

& InstructGPT (davinci-003) & $+$ &  & $\sim$ & $\sim$ & $--$ & $--$ & $--$ \\

& InstructGPT (turbo) & $++$ & $\sim$ &  & $\sim$ & $--$ & $--$ & $--$ \\

& ChatGPT & $\sim$ & $\sim$ & $\sim$ &  & $--$ & $--$ & $--$ \\

& PaLM-2 (text-bison) & $++$ & $++$ & $++$ & $++$ &  & $\sim$ & $--$ \\

& GPT4 & $++$ & $++$ & $++$ & $++$ & $\sim$ &  & $-$ \\

& Falcon (180B-Chat) & $++$ & $++$ & $++$ & $++$ & $++$ & $+$ &  \\

\midrule
\multicolumn{2}{l}{uncertainty} \\
& InstructGPT (davinci-002) &  & $\sim$ & $\sim$ & $\sim$ & $\sim$ & $\sim$ & $--$ \\

& InstructGPT (davinci-003) & $\sim$ &  & $\sim$ & $\sim$ & $\sim$ & $\sim$ & $--$ \\

& InstructGPT (turbo) & $\sim$ & $\sim$ &  & $\sim$ & $\sim$ & $\sim$ & $--$ \\

& ChatGPT & $\sim$ & $\sim$ & $\sim$ &  & $--$ & $--$ & $--$ \\

& PaLM-2 (text-bison) & $\sim$ & $\sim$ & $\sim$ & $++$ &  & $\sim$ & $--$ \\

& GPT4 & $\sim$ & $\sim$ & $\sim$ & $++$ & $\sim$ &  & $--$ \\

& Falcon (180B-Chat) & $++$ & $++$ & $++$ & $++$ & $++$ & $++$ &  \\

\bottomrule
\end{tabular}
    \caption{Epistemological dimensions. Pairwise t-test results. $--$ and $++$ indicate high significance ($p<0.01$) that the model in the row scores lower/higher than the model in the colum. Single $-/+$ indicate $p < 0.05$ and $\sim$ indicates no significant difference between the models.}
    \label{tab:pairwise_epistemological}
\end{table}

\begin{table}[htp]
    \centering
    \footnotesize
\resizebox{\textwidth}{!}{%
    \begin{tabular}{lSSSSSSSS}
\toprule
Issue & \multicolumn{3}{c}{InstructGPT}  & \multicolumn{1}{c}{ChatGPT} & \multicolumn{1}{c}{PaLM-2} & \multicolumn{1}{c}{GPT4} & \multicolumn{1}{c}{Falcon}  \\
& \multicolumn{1}{c}{\scriptsize davinci-002} & \multicolumn{1}{c}{\scriptsize davinci-003} & \multicolumn{1}{c}{\scriptsize turbo} & \multicolumn{1}{c}{} & \multicolumn{1}{c}{\scriptsize text-bison} & \multicolumn{1}{c}{} & \multicolumn{1}{c}{\scriptsize 180B-Chat}  \\
\midrule
\textbf{style} \\
inconsistent & 4.88 & 1.00 & 0.33 & 0.00 & 1.11 & 0.22 & 0.45 \\
repetitive & 20.15 & 3.11 & 0.11 & 0.56 & 1.45 & 1.11 & 0.33 \\
too informal & 4.11 & 1.11 & 0.22 & 0.11 & 1.78 & 1.44 & 0.89 \\
too long & 1.03 & 1.67 & 0.33 & 0.89 & 2.12 & 2.11 & 0.89 \\
too short & 10.14 & 8.56 & 0.22 & 0.22 & 2.56 & 0.33 & 1.11 \\
\vspace{5pt}other & 2.95 & 1.00 & 0.22 & 0.00 & 0.78 & 0.67 & 0.45 \\

\textbf{clarity} \\
hard math & 1.67 & 0.44 & 1.67 & 0.33 & 0.67 & 1.56 & 0.00 \\
sentences too long & 1.80 & 1.33 & 0.11 & 0.22 & 1.67 & 3.11 & 1.22 \\
too technical & 3.59 & 1.00 & 0.33 & 0.44 & 1.22 & 2.56 & 0.56 \\
\vspace{5pt}other & 8.60 & 1.00 & 0.33 & 0.11 & 1.56 & 0.44 & 0.78 \\

\textbf{correctness} \\
incomplete sentence & 3.47 & 2.44 & 0.78 & 0.00 & 0.00 & 0.00 & 0.00 \\
incorrect grammar & 6.29 & 0.33 & 0.33 & 0.11 & 0.11 & 0.11 & 0.67 \\
incorrect punctuation & 2.18 & 0.44 & 0.56 & 0.00 & 0.11 & 0.22 & 0.56 \\
incorrect spelling & 0.77 & 0.00 & 0.11 & 0.11 & 0.22 & 0.00 & 0.11 \\
\vspace{5pt} other & 3.98 & 1.11 & 0.00 & 0.11 & 0.45 & 0.11 & 0.78 \\

\textbf{tone} \\
biased & 28.50 & 34.44 & 24.78 & 42.38 & 33.85 & 30.33 & 23.72 \\
negative & 1.28 & 1.78 & 1.56 & 1.00 & 5.01 & 3.00 & 1.89 \\
persuasive & 2.57 & 8.00 & 4.80 & 7.68 & 10.69 & 8.00 & 4.45 \\
\vspace{5pt}other & 0.39 & 0.67 & 0.22 & 0.11 & 0.45 & 2.00 & 0.22 \\

\textbf{accuracy} \\
anecdotal & 10.78 & 1.33 & 3.35 & 19.24 & 5.90 & 2.56 & 3.01 \\
incorrect & 20.92 & 10.78 & 5.58 & 3.23 & 11.69 & 4.44 & 4.57 \\
science out of context & 9.37 & 6.11 & 5.69 & 2.67 & 5.35 & 3.78 & 2.56 \\
self contradictory & 2.70 & 0.89 & 0.11 & 0.11 & 0.89 & 0.44 & 0.33 \\
wrong use of terms & 1.93 & 1.22 & 1.00 & 0.33 & 1.45 & 0.44 & 0.22 \\
\vspace{5pt} other & 3.34 & 3.00 & 1.90 & 1.89 & 2.00 & 5.67 & 0.78  \\

\textbf{specificity} \\
irrelevant info & 15.15 & 4.56 & 3.79 & 5.12 & 8.69 & 8.89 & 4.01 \\
vague & 49.42 & 44.78 & 48.88 & 58.40 & 51.67 & 39.11 & 35.86 \\
\vspace{5pt}other & 1.67 & 3.44 & 1.45 & 0.56 & 2.12 & 1.67 & 1.34 \\

\textbf{completeness} \\
does not address main parts & 29.91 & 22.56 & 11.16 & 9.79 & 15.92 & 8.78 & 9.47 \\
does not address region & 3.34 & 2.67 & 0.78 & 0.56 & 1.34 & 1.78 & 1.22 \\
does not address time & 2.05 & 4.11 & 1.90 & 0.67 & 0.67 & 2.78 & 0.67 \\
ignores science & 9.11 & 14.11 & 6.92 & 5.01 & 10.47 & 5.44 & 3.01 \\
not enough detail & 52.89 & 60.22 & 64.06 & 79.53 & 58.13 & 61.22 & 51.89 \\
\vspace{5pt} other & 1.16 & 0.89 & 0.45 & 0.11 & 0.89 & 2.78 & 1.11 \\

\textbf{uncertainty} \\
consensus missing & 19.77 & 14.89 & 21.99 & 9.34 & 12.14 & 9.89 & 9.80 \\
contradicting evidence missing & 4.11 & 6.33 & 2.57 & 2.00 & 4.23 & 3.56 & 2.90 \\
uncertainty missing & 57.25 & 75.00 & 72.88 & 87.65 & 71.94 & 76.78 & 58.02 \\
other & 0.90 & 1.11 & 0.45 & 0.11 & 0.45 & 1.89 & 0.33 \\
\bottomrule
    \end{tabular}%
}
    \caption{Percentage of specific issues identified by raters.}
    \label{tab:issue_frequencies}
\end{table}

\newpage
\subsection{Questions}
In this section we explain the pipeline used for selection, generation, post-processing and sampling climate change related questions. The question set consists of $300$ questions, with 100 questions gathered from $3$ sources each: i) Synthetic questions generated based on Wikipedia articles, ii) Manually rephrased questions based on Skeptical Science website, and iii) questions taken from Google Trends. 
\subsubsection{Synthetic Questions from Wikipedia} \label{sec:synthetic-questions}
We started by gathering a set of Wikipedia articles related to climate change. We followed $3$ strategies to select climate related articles from Wikipedia. Following the first strategy (\refer), we gather all the Wikipedia articles that are referenced in the main ``Climate Change'' article.\footnote{\url{https://en.wikipedia.org/wiki/Climate_change}} In the second strategy (\categ), we select all the articles that are directly listed in the climate change category. Finally, to cover regional articles (\regio), we manually curate a list of articles with titles \textit{``Climate Change in [country/region]''}. From a pool of articles gathered following these $3$ strategies, we selected paragraphs within an article if the paragraph consists of more than $500$ characters. In total, we obtained $1969$ paragraphs from Wikipedia. The following table reports a break-down of number of paragraphs based on the selection strategy: %

\begin{center}
\begin{tabular}{lcc}\toprule
Strategy &  \# Articles &  \# Paragraphs \\ \midrule
\refer{} & $35$ & $858$ \\
\categ{} & $46$ & $434$ \\
\regio{} & $48$ & $677$ \\ \midrule
Total & $129$ & $1969$ \\ \bottomrule
\end{tabular}
\end{center}

We then input each selected paragraph in \gptfour{}. We ask the model to generate as many questions as possible that can be answered using the paragraph. The model is instructed to only generate questions that are salient and related to climate change. This process resulted in $15265$ questions. We post process the questions and remove undesirable ones with $4$ filters that we explain next.

\paragraph{Climate Change Filter.} We remove all questions that are not climate change related. We use the climate-bert \citep{climatebert} classifier and label each question with two labels: climate related and not climate related. We remove $2647$ questions that are not classified as climate-related questions. 

\paragraph{Duplicate Filter.} We remove questions that are a duplicate of another question. To this end, we embed all questions using a universal sentence encoder.%
\footnote{We use \texttt{universal-sentence-encoder-qa/3} model.} We consider two questions as duplicates if the cosine similarity between their embeddings is greater than $0.85$. Therefore, we remove $1188$ questions that are duplicates of other questions. 

\paragraph{Context Dependent Filter.} We filter out questions that are taken out of context.  The reason that this filter is necessary is that we generate questions from paragraphs, therefore, some questions are nonsensical when they are not accompanied by the corresponding Wikipedia paragraph. An example of such a question is: \emph{“What are the two classes of climate engineering discussed in the study?”};  without knowing which study is referred to, this question cannot be answered. To develop this filter, we build a dedicated classifier using in-context probing \citep{icp}.
Specifically, we manually annotate $100$ questions with two labels: context dependent, and not context dependent. Next, we contextualize the question with the instruction \textit{``Write Yes if the query is taken out of context, write No otherwise.''} and extract the last layer's representations of a \texttt{flan-xxl} encoder~\citep{chung2022scaling}. Finally, we train a logistic regression probing classifier on the representations to detect context dependent questions. We find the context dependency filter to be $97\%$ accurate on $100$ manually annotated validation questions. Using this classifier, we detect $552$ context dependent questions. 

\paragraph{Specificity Filter.} We remove questions that are asking about a very specific and narrow topic. In our study, we aim to evaluate large language models on a set of challenging and multifaceted questions that target information needs of users related to climate change. Therefore, questions that ask for a specific detail are not the target of this study and are typically easy to answer. An example of such question is: ``What was the reason for shutting down reactor number one of the Fessenheim Nuclear Power Plant on 4 August 2018?'' To remove such specific questions, we again build in-context probing classifier on top of \texttt{flan-xxl} representations. We contextualize each question with the instruction: \textit{``Write Yes if the following query is asking about a specific subject, write No otherwise''.} and train the probe on top of extracted contextualized representations from the last layer of \texttt{flan-xxl}. We find the specificity filter to be $84\%$ accurate on a sample of $100$ annotated validation questions. We detect and remove $5472$ specific questions.

After applying all $4$ filters, the final post-processed question set consists of $5404$ questions. The question set that is rated in our evaluation framework consists of $100$ questions from each source. This means that we need to sample $100$ diverse questions from this pool of $\approx 5$k questions. To make sure that we cover different topics and type of questions, we first label each question with the topic and properties of the question, and then sample a $100$ validation questions, where different topics and properties are equally presented. Next, we explain the classifiers that are developed for labeling the questions.

\begin{figure}
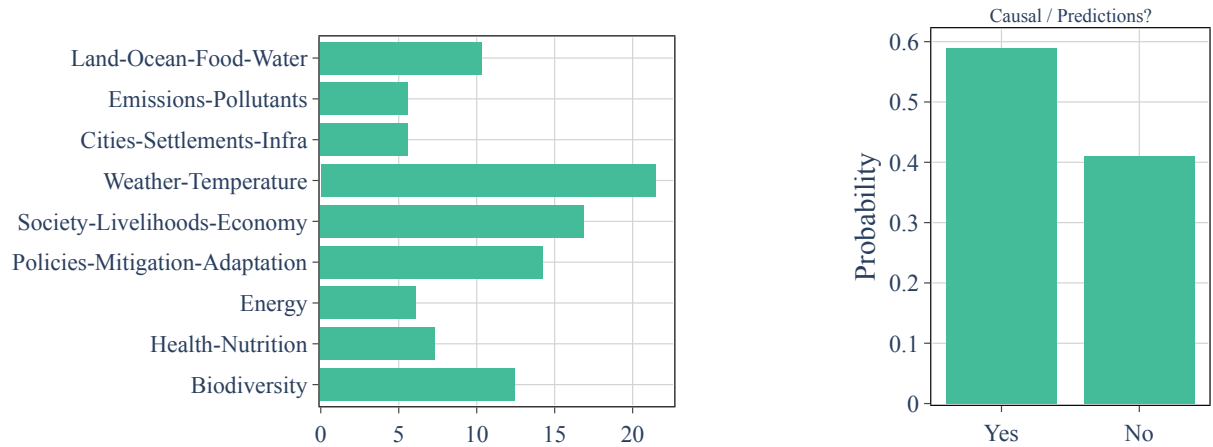

    \centering
    \begin{subfigure}[b]{0.55\textwidth}
        \centering
        \includesvg[width=\textwidth]{figures/topics.svg}
    \end{subfigure}
    \hfill
    \begin{subfigure}[b]{0.35\textwidth}
        \centering
        \includesvg[width=0.8\textwidth]{figures/causal.svg}
    \end{subfigure}
    \caption{Left: Distribution of the topic of the synthetic questions. The x axis shows the percentage of questions in each topic. The most frequent topic is ``Weather-Temperature'' topic. Right: Distribution of the causal feature of synthetic questions. There are more questions that are asking causal questions, or questions about predicting the future compare to the rest of the questions.}
    \label{fig:qdist}
\end{figure}

\paragraph{Topic Classifier.} We use the same in-context probing 
approach as above and train a logistic regression classifier on top of \texttt{flan-xxl} encoder to classify questions based on the topics. Inspired by IPCC chapters, we consider the following $9$ topics: \textit{``Energy'',
 ``Emissions-Pollutants'',
 ``Policies-Mitigation-Adaptation'',
 ``Weather-Temperature'',
 ``Land-Ocean-Food-Water'',
 ``Society-Livelihoods-Economy'',
 ``Health-Nutrition'',
 ``Biodiversity'',
 ``Cities-Settlements-Infra''}. We find this classifier to be $75\%$ accurate on a sample of $100$ annotated validation questions. The distribution of predicted questions' topics is depicted in \cref{fig:qdist}.
 
\paragraph{Causal Prediction Classifier.} An important and challenging type of questions that one can ask about climate change is about causes or effects of climate change, or predictions about the future. To detect this type of questions, we classify questions into two classes: causal-prediction class and others. The instruction that is used for contextualizing the questions is: \textit{``Write Yes if the following query is asking about causes or effects of something, or is asking about predictions about the future. write No otherwise''}. We find this classifier to be $93\%$ accurate on a sample of $100$ annotated validation questions. The distribution of predictions is shown in \cref{fig:qdist}

For synthetic Wikipedia questions, we sample $108$ questions, $6$ from each of the $18$ possible combinations of topic and causal-prediction features. We sample an equal number of questions from the other two sources, Skeptical Science and GTrends. We manually inspect all questions from all the $3$ different sources. The final question set used in the evaluations consists of $300$ questions, i.e., $100$ questions from each source.

\newpage
\subsection{Prompts}
Please refer to \cref{tab:prompts} for an extensive list of prompts used to generate the data used throughout the paper.
\begin{table}[p]
\centering
\scriptsize
\resizebox{\textwidth}{!}{%
\begin{tabular}{@{}p{3.5cm}p{9.5cm}@{}}\toprule
         Function & Prompt  \\ \midrule
Answer Generation      &  You are an expert on climate change communication. Answer each question in a 3-4 sentence paragraph.\\[0.4cm]
Obtain URL      &  Please provide a Wikipedia article that supports your answer. Just state the url, do not include additional text. If there is no Wikipedia url supporting the answer just say ``No URL''.\\ [0.4cm]
Extract Keypoints  & Now go through all the statements made in the answer. Mention 1 to 3 key statements that are made to answer the question. If you can not provide key statement/statements, only write No Keypoints. It is very important to copy the statements verbatim from the answer. \\[0.7cm]
Rate Passages & You are given a statement\tablefootnote{We found that we have used \emph{``an statement''} instead of \emph{``a statement''} in our experiments. %
We did not rerun the experiments as we believe that LLMs are quite robust to minor typos and results should not be significantly affected by it. Also, doing otherwise would be wasteful.} and a passage from Wikipedia. Rate how useful the passage is for evaluating the statement on a scale from 0 (completely irrelevant) to 100 (supports or contradicts the statement). Rate the passage high only if it supports or contradicts the statement. Just state the numbers in one line, nothing else. Statement: [keypoint] Passage: [par] \\\midrule
Presentational AI~Assistance & Given the following question and answer, express your disagreement with the statement in a concise sentence in a single line. You may be provided with relevant paragraphs from Wikipedia, if so, you must use those verbatim to support your critique. If you fully agree with the statement, state ``No Critique''. Question: [question] Answer: [answer] Statement: [statement] \\[1.0cm]
Style Statement & The information is presented well for a general audience. In particular, the answer is not too long or too short, there is no repetition in the text, and the answer is not too informal or too technical. \\ [0.4cm]
Clarity Statement & The answer is clear and easy to understand. For example, if there are numbers and formulae in the answer, they are easy to understand. Furthermore, sentences are not too long or too short. \\ [0.4cm]
Correctness Statement & The language in the answer does not contain mistakes. In particular, there are no grammatical, spelling, or punctuation errors. \\ [0.4cm]
Tone Statement & The tone of the answer is neutral and unbiased. In particular, the tone is not negative and the answer does not try to convince the reader of an opinion or belief. \\ \midrule
Epistemological AI~Assistance & Given the following question and answer, express your disagreement with the statement in a concise sentence in a single line. You may be provided with relevant paragraphs from Wikipedia, if so, you must use those verbatim to support your critique. If you fully agree with the statement, state ``No Critique''. Question: [question] Answer: [answer] Statement: [statement].\\[0.9cm]
Accuracy Statement & The answer is accurate. In particular, it does not take scientific findings out of context, does not contradict itself, does not rely on anecdotal evidence, and does not misuse key terms or scientific terminology.\\[0.6cm]
Specificity Statement & There is no irrelevant statement with respect to the question in the answer, and there is no vague or generic statement in the answer.\\[0.4cm]
Completeness Statement & The answer addresses everything the question asks for. In particular, it does not miss any part of the question and provides enough necessary details, e.g., numbers, statistics, and details. If the question asks for a specific time range or region, the answer correctly provides that information. \\[0.6cm]
Uncertainty Statement & If there is an uncertainty involved in the scientific community, the answer appropriately conveys that uncertainty. Note that it may be appropriate not to mention uncertainty at all.\\
\midrule
Dimension-aware main prompt& You are an expert on climate change communication. Answer the question in a 3-4 sentence paragraph. The answer should be concise and tailored for a general audience. It must be clear, and easy to understand. The answer should be presented in a neutral, unbiased tone without any negative connotations or attempts to persuade. The answer should be factually accurate. The answer should be specific to the question and avoid irrelevant, generic, or vague statements. The answer should comprehensively address all aspects of the question. Where scientific uncertainty exists, the answer should appropriately reflect this, conveying the range of scientific perspectives or the limitations of current knowledge.\\
\bottomrule
\end{tabular}%
}
\caption{Prompts used to generate answers, AI Assistance and evidence.}\label{tab:prompts}
\end{table}

\newpage
\subsection{Answer Statistics} \label{sec:answer-stats}
We report the average number of sentences and the average number of words per sentence for all models evaluated in \cref{tab:answer-stats}. Although in the prompts used for answer generation we explicitly instruct the model to only use $3$ to $4$ sentences to answer the question, we observe that most models generate between $4$ to $5$ sentences. Furthermore, \gptturbo{}, \gptfour{}, and \falcon{}, generate longer sentences compared to the other models.
\begin{table}[htbp]
\centering
\scriptsize
\resizebox{\textwidth}{!}{%
\begin{tabular}{lSSSSSSSS}
\toprule
 & \multicolumn{3}{c}{InstructGPT}  & \multicolumn{1}{c}{ChatGPT} & \multicolumn{1}{c}{PaLM-2} & \multicolumn{1}{c}{GPT4} & \multicolumn{1}{c}{Falcon}  \\
& \multicolumn{1}{c}{\scriptsize davinci-002} & \multicolumn{1}{c}{\scriptsize davinci-003} & \multicolumn{1}{c}{\scriptsize turbo} & \multicolumn{1}{c}{} & \multicolumn{1}{c}{\scriptsize text-bison} & \multicolumn{1}{c}{} & \multicolumn{1}{c}{\scriptsize 180B-Chat}  \\
\midrule
\# Sentences & 4.99 & 3.11 & 3.42 & 4.07 & 4.47 & 4.33 & 3.81 \\
\# Words per sentence & 14.3 & 18.68 & 21.49 & 20.66 & 19.67 & 21.52 & 22.03 \\ \bottomrule
\end{tabular}%
}
\caption{Average number of sentences and words per sentence for each model. We observe $4$ out of $7$ models generate $4$ to $5$ sentences, and \falcon{} generates longer sentences compared to the other models in the batch.}\label{tab:answer-stats}
\end{table}

\newpage
\subsection{Rating Framework Details}
\label{sect:appendix-rating-framework}

\subsection{Rater Demographics}
\label{sect:appendix-raters-demographics}

We are working with a group of $32$ raters. The raters are all fluent in English and all have at least an undergraduate degree in a climate-related field of study. This includes environmental disciplines (e.g. environmental science, earth science, atmospheric physics, ecology, environmental policy, climate economics), and also other disciplines (including the behavioral and social sciences) as long as their academic work (coursework, project work, or otherwise) involves work on climate or environmental studies. The remaining demographics can be seen in \cref{tab:dem}.

\begin{table}
\begin{subtable}[b]{0.45\textwidth}
\centering
\begin{tabular}[b]{lr}
\toprule
Age bracket & $\%$ \\
\midrule
$[18, 25)$ & $43.75$ \\
$[25, 35)$ & $34.38$ \\
$[35, 45)$ & $12.50$ \\
$[45, 55)$ & $6.25$ \\
$[55, 65)$ & $3.12$ \\
\bottomrule
\end{tabular}
\caption{Distribution of age of our raters.}\label{tab:dem-age}
\end{subtable}
\hfill
\begin{subtable}[b]{0.45\textwidth}
\centering
\begin{tabular}[b]{lr}
\toprule
Sex & $\%$ \\
\midrule
Female & $56.25$ \\
Male & $43.75$ \\
\bottomrule
\end{tabular}
\caption{Distribution of sex of our raters.}\label{tab:dem-sex}
\end{subtable}
\\
\\
\begin{subtable}[b]{0.45\textwidth}
\centering
\begin{tabular}[b]{lr}
\toprule
Ethnicity & $\%$ \\
\midrule
White & $68.75$ \\
Black & $12.50$ \\
Asian & $12.50$ \\
Mixed & $3.12$ \\
Other & $3.12$ \\
\bottomrule
\end{tabular}
\caption{Distribution of simplified ethnicities of our raters.}\label{tab:dem-eth}
\end{subtable}
\hfill
\begin{subtable}[b]{0.45\textwidth}
\centering
\begin{tabular}[b]{lr}
\toprule
Country of residence & $\%$ \\
\midrule
United Kingdom & $25.00$ \\
South Africa & $12.50$ \\
Portugal & $12.50$ \\
United States & $9.38$ \\
Greece & $6.25$ \\
New Zealand & $6.25$ \\
Netherlands & $6.25$ \\
Poland & $6.25$ \\
Canada & $3.12$ \\
Germany & $3.12$ \\
Czech Republic & $3.12$ \\
Hungary & $3.12$ \\
Italy & $3.12$ \\
\bottomrule
\end{tabular}
\caption{Distribution of countries of residence of our raters.}\label{tab:dem-res}
\end{subtable}
\caption{Demographic information of our raters.}\label{tab:dem}
\end{table}

\subsection{Rating Statements}
\begin{table}[htp]
    \centering
        \small
    \begin{tabular}{lp{9cm}}\toprule

\textbf{Presentational Dimensions} & \textbf{Statement and possible issues}\\

\midrule

style & The information is presented well (for a general audience).\\
\hskip 10pt too informal & \hskip 10pt $\square$ too informal/colloquial \\
\hskip 10pt too long & \hskip 10pt $\square$ answer too long \\
\hskip 10pt too short& \hskip 10pt $\square$ answer too short \\
\hskip 10pt inconsistent & \hskip 10pt $\square$ inconsistent language/style/terminology \\
\hskip 10pt repetitive & \hskip 10pt $\square$ repetitive \\  %
\hskip 10pt other & \hskip 10pt $\square$ other \vspace{5pt} \\

clarity & The answer is clear and easy to understand.\\
\hskip 10pt sentences too long & \hskip 10pt $\square$ sentences too long \\ 
\hskip 10pt too technical & \hskip 10pt $\square$ language too technical \\ 
\hskip 10pt hard math & \hskip 10pt $\square$ numbers/formulae hard to understand \\
\hskip 10pt other & \hskip 10pt $\square$ other \vspace{5pt} \\

correctness & The language in the answer does not contain mistakes.\\
\hskip 10pt incomplete sentence & \hskip 10pt $\square$ sentence is incomplete \\
\hskip 10pt incorrect spelling & \hskip 10pt $\square$ spelling mistakes \\
\hskip 10pt punctuation mistakes& \hskip 10pt $\square$ punctuation mistakes \\
\hskip 10pt incorrect grammar & \hskip 10pt $\square$ grammatical errors \\
\hskip 10pt other & \hskip 10pt $\square$ other \vspace{5pt} \\

tone & The tone of the answer is neutral and unbiased.\\
\hskip 10pt biased & \hskip 10pt $\square$  the answer is biased\\
\hskip 10pt persuasive & \hskip 10pt $\square$  tries to convince me of an opinion/belief\\
\hskip 10pt negative & \hskip 10pt $\square$  the tone is too negative\\
\hskip 10pt other & \hskip 10pt $\square$ other \vspace{5pt} \\

\midrule

\textbf{Epistemological Dimensions} & \\ %

\midrule

 accuracy & The answer is accurate.\\
\hskip 10pt incorrect & \hskip 10pt $\square$  incorrect\\
\hskip 10pt science out of context & \hskip 10pt $\square$  takes scientific findings out of context\\
\hskip 10pt self contradictory & \hskip 10pt $\square$  self-contradictory\\
\hskip 10pt wrong use of terms & \hskip 10pt $\square$  wrong use of key terms/scientific terminology\\
\hskip 10pt other & \hskip 10pt $\square$ other \vspace{5pt}\\

 specificity & The answer addresses only what the question asks for, without adding irrelevant information.\\
\hskip 10pt irrelevant info & \hskip 10pt $\square$  includes irrelevant parts\\
\hskip 10pt vague & \hskip 10pt $\square$  too vague/unspecific\\
\hskip 10pt other & \hskip 10pt $\square$ other \vspace{5pt}\\

 completeness & The answer addresses everything the question asks for.\\
\hskip 10pt does not address main parts & \hskip 10pt $\square$  misses important parts of the answer\\
\hskip 10pt does not address region & \hskip 10pt $\square$  does not address the region the question asks about\\
\hskip 10pt does not address time & \hskip 10pt $\square$  does not address time or time range the question asks about\\
\hskip 10pt not enough detail & \hskip 10pt $\square$  does not give enough detail (e.g. numbers, statistics, details)\\
\hskip 10pt ignores science & \hskip 10pt $\square$  ignores relevant scientific knowledge \\
\hskip 10pt other & \hskip 10pt $\square$ other \vspace{5pt}\\

 uncertainty & The answer appropriately conveys the uncertainty involved.\\
\hskip 10pt uncertainty missing & \hskip 10pt $\square$  degree of (un)certainty not given when it should be\\
\hskip 10pt consensus missing & \hskip 10pt $\square$  agreement in the scientific community not given when important\\
\hskip 10pt contradicting evidence missing & \hskip 10pt $\square$  contradicting evidence (if existing) not mentioned\\
\hskip 10pt other & \hskip 10pt $\square$ other \vspace{5pt}\\
\bottomrule
    \end{tabular}
    \caption{(on the right) Statements as presented to the raters. We query each dimension separately in the interface (\cref{fig:interface_cow_rated}) and ask \emph{"To what extent do you agree with the statement below?"} We also require the raters to identify particular issues for the given list if they disagree with a statement. On the left side we list the dimensions the statements belong in and a shorthand for the issue names used in tables throughout this work.}
    \label{tab:eval_statements}
\end{table}

For presentational and epistemological accuracy we evaluate 4 dimensions each. Given a question-answer pair the raters are asked to what degree they agree with one of the statements in \cref{tab:eval_statements}.\footnote{Please note that when we use the shorthand \emph{correctness} in our results, this only refers to correctness \emph{of the language}, i.e.\ presentational correctness. The corresponding epistemological dimension is \emph{accuracy}, i.e.\ correctness \emph{of the answer}.} The raters select agreement on a 5-point scale from \emph{completely disagree} to \emph{completely agree}. For the two lowest choices we ask for additional details which can be selected from a list of possible issues, including \emph{other} which allows free-text input. See \cref{sect:appendix-template-screenshots} for screenshots of the rating interface.

\subsection{Tutorial and Admission Test}
\label{sect:appendix-tutorial-and-admission}
We devise a special introduction session for new participants that contains a tutorial followed by an admission test. The purpose of the session is twofold:
(1) The introduction session is designed to familiarize the raters with the interface and the task.
(2) Based on the session's outcome we select raters into the rating pool.

\begin{figure}[htp]
    \centering
    \includegraphics[width=\textwidth]{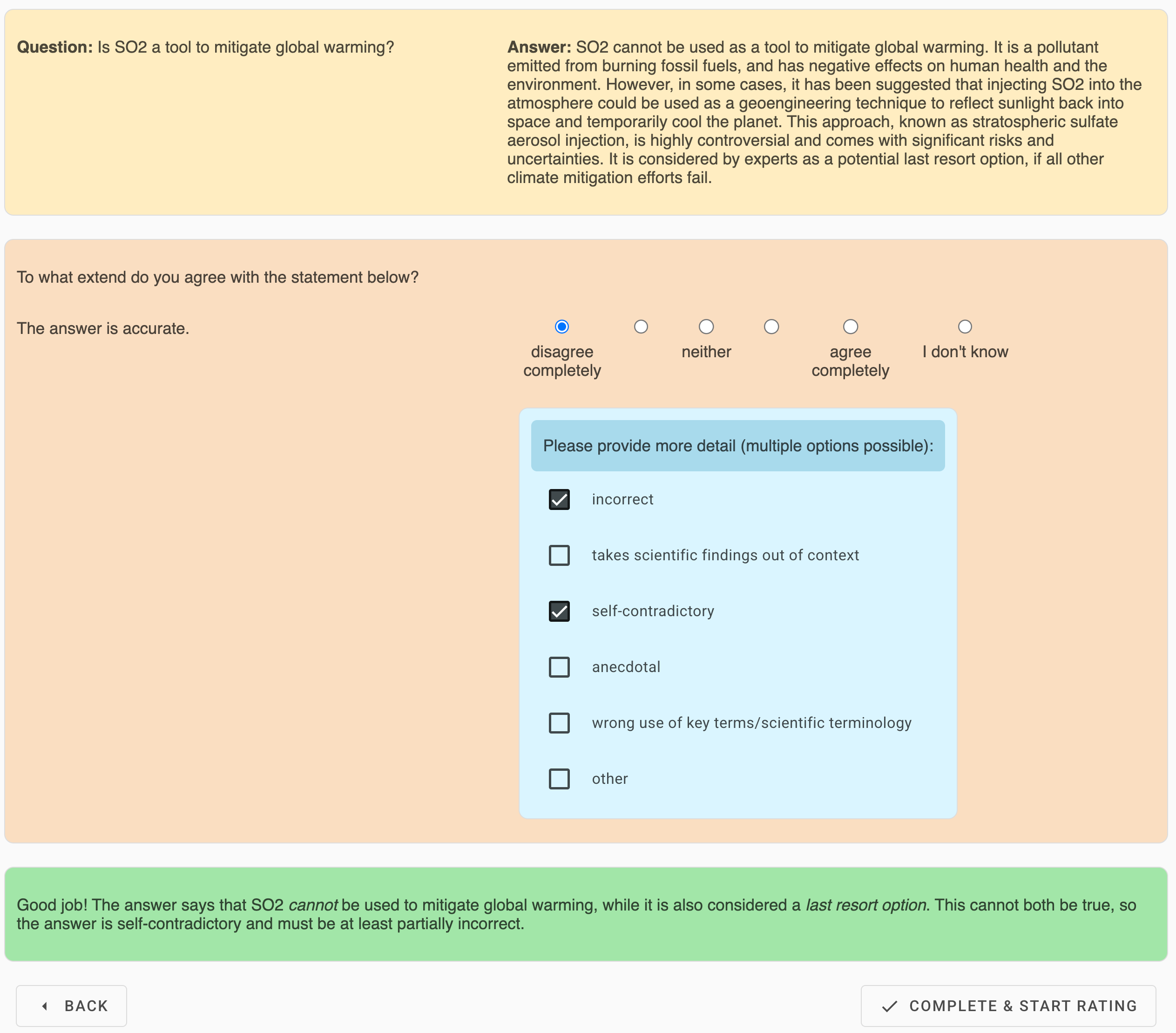}
    \caption{Screenshot of the last of 4 tutorial questions with the correct answer selected.}
    \label{fig:screenshot_tutorial}
\end{figure}

\textbf{Tutorial} \hskip 5pt
In the tutorial (see \cref{fig:screenshot_tutorial}) we present 4 examples of increasing difficulty in the rating interface and only ask for one dimension each. Each example exhibits a particular main issue and we expect raters to identify this issue correctly. A hint is given if the rater selects a wrong answer that does not identify the issue and they can only proceed to the next item if an acceptable answer that does identify the issue is given. Regarding other issues than the main issue, one might disagree on some of these issues and we allow several possible selections. Note that to identify the main issue, a low rating (disagree completely or disagree) must be selected. Once a valid response is selected we show positive feedback and explain why the outcome is the desired one. We don't collect any data during the tutorial part.

\textbf{Admission Test} \hskip 5pt
To test the raters' ability and attention to detail we select three realistic examples that exhibit at least one major flaw. We use the full template and ask about all statements in \cref{tab:eval_statements}. We record the responses and assign or deduct points for every detected, undetected, and over-detected issue. The point scheme was decided among the authors after carefully considering possible disagreements or subjective interpretations.

Based on the performance of an early group of raters with known performance on the task we decide on a threshold and admit raters above that score to the pool. We believe that the tutorial and admission test were effective in ensuring that raters were both familiar with the interface as well as the type of assessment we expect from them, which requires close reading of question and answer, basic knowledge of climate change, and an understanding of the tasks dimensions and issues that allows them to rate dimensions and select specific issues reliably.

\subsection{Template Screenshots}
\label{sect:appendix-template-screenshots}
\begin{figure}
    \centering
    \includegraphics[width=\textwidth]{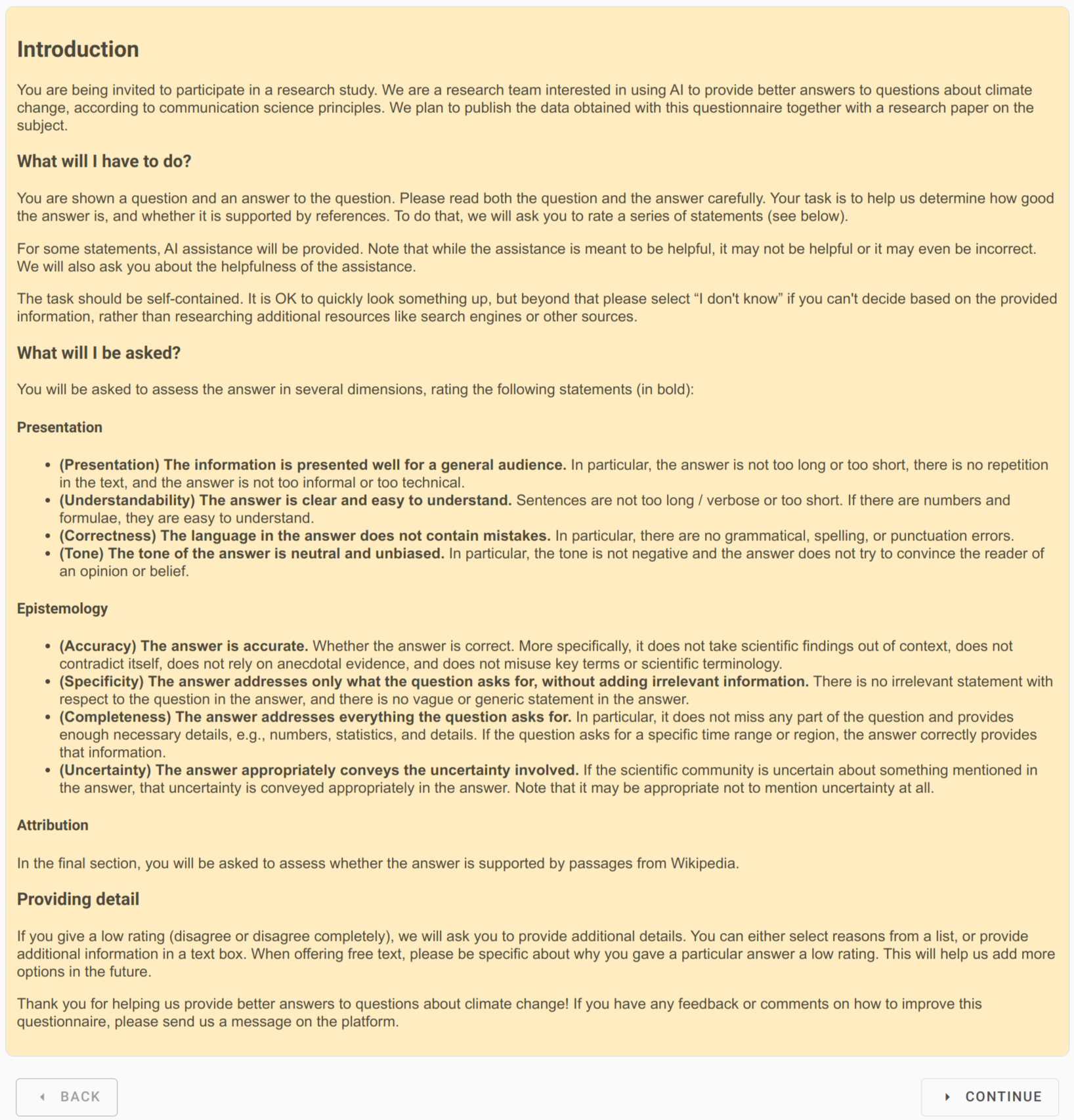}
    \caption{Screenshot of the instructions to the raters, provided at the beginning of the first rating session.}
    \label{fig:screenshot_instructions}
\end{figure}

\begin{figure}
    \centering
    \includegraphics[width=\textwidth]{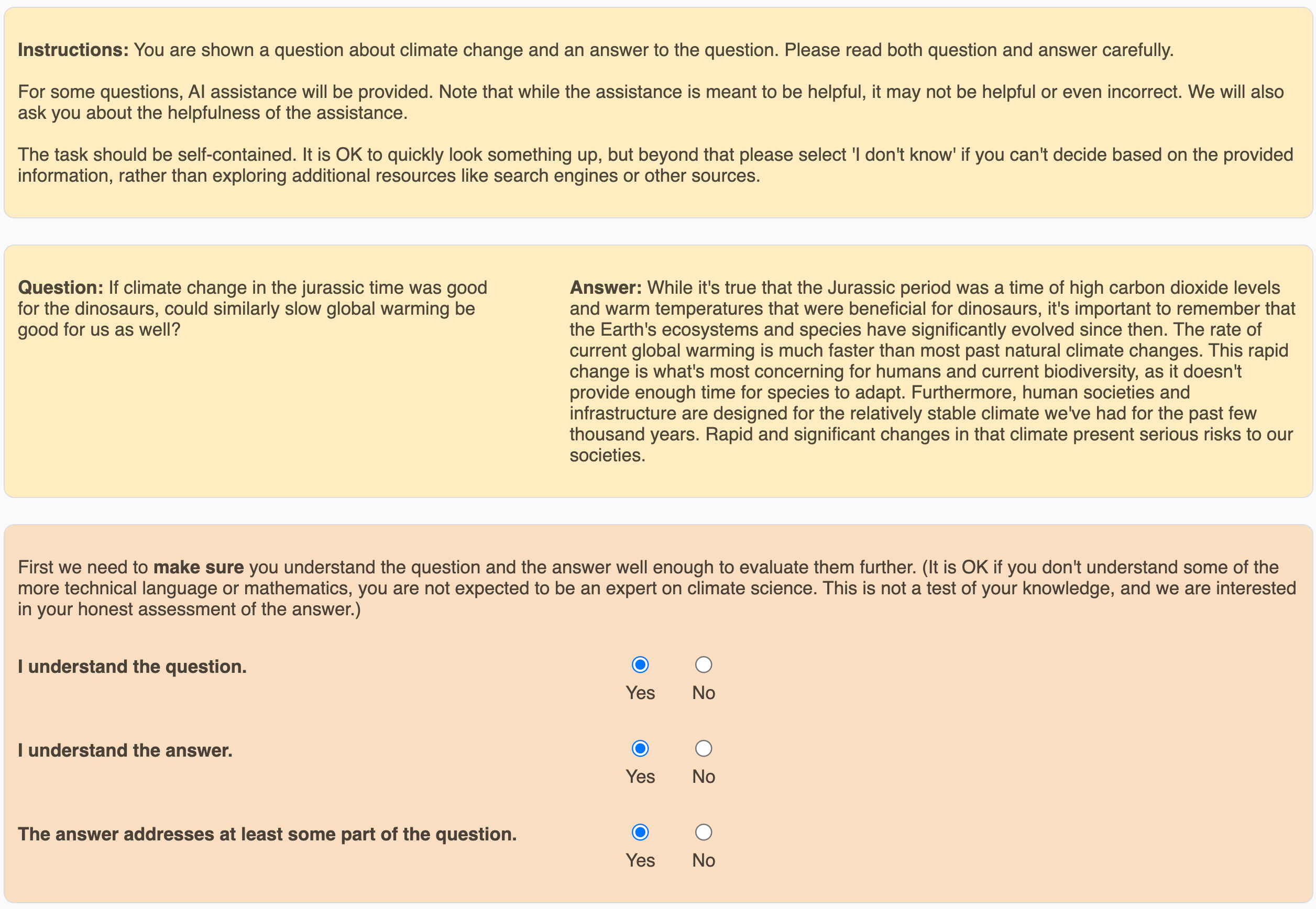}
    \caption{Screenshot of screening template which ensures the raters are able to judge the example. If any of the questions is answered \emph{No} we skip the example and proceed with the next one.}
    \label{fig:screenshot_screening_noais}
\end{figure}

\begin{figure}
    \centering
    \includegraphics[width=\textwidth]{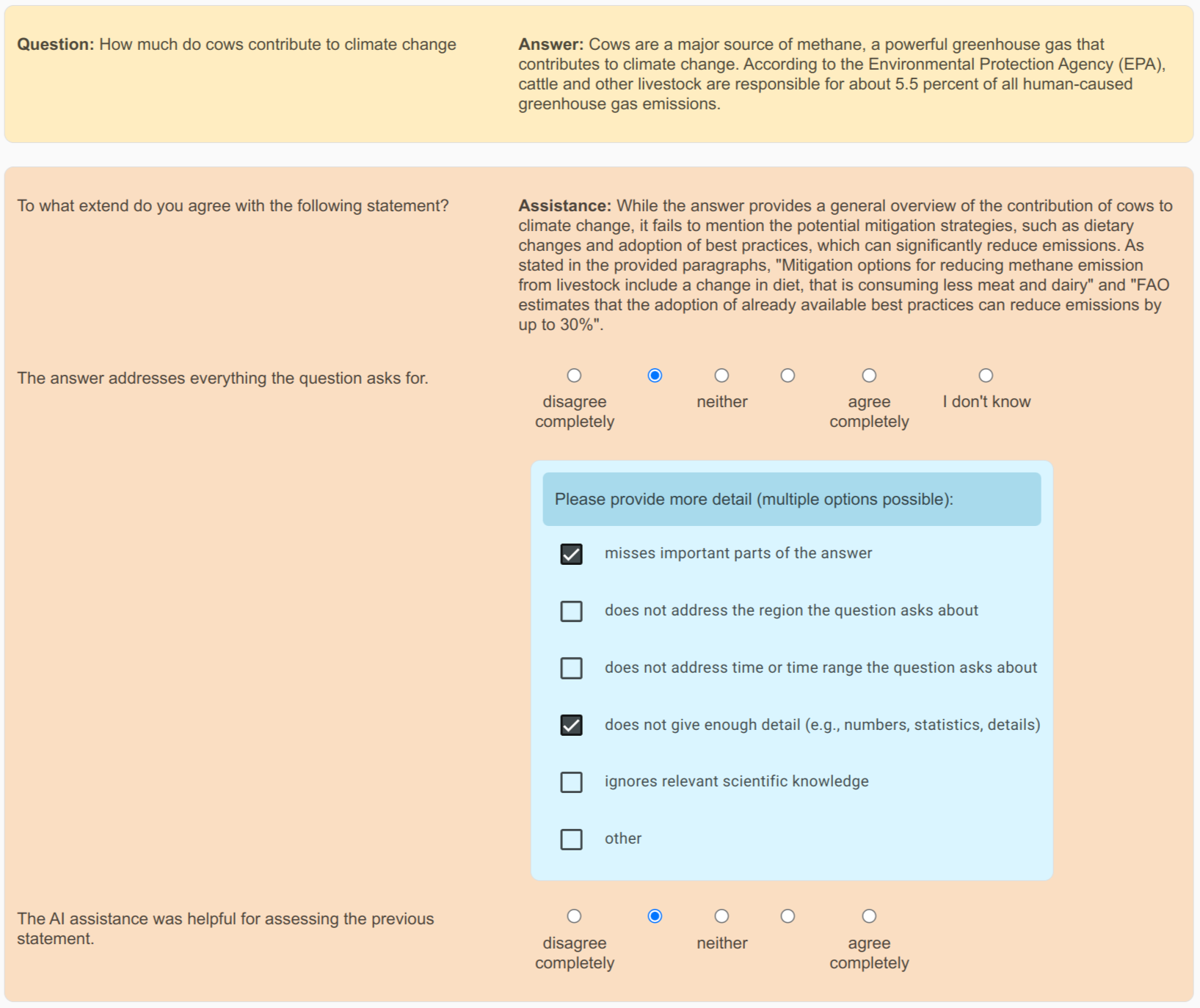}
    \caption{Screenshot of the rating interface. Here we are rating the \emph{completeness} dimension. Since the rater selected a low rating, the interface asks for specific details. For ratings of \emph{3 = neither} and above, the blue box is hidden. In this example we also show the AI Assistance which is not always available, which is rated as not helpful.}
    \label{fig:interface_cow_rated}
\end{figure}

We provide screenshots of our rating implementation in \cref{fig:screenshot_instructions,fig:screenshot_screening_noais,fig:interface_cow_rated}.

\newpage
\subsection{Inter-Rater Agreement} \label{sec:agreement}
We first measure the agreement among raters when rating each dimension on the likert scale. In particular, we report two metrics of agreement:

\textbf{Pairwise distance.} We measure the average pairwise distance between the ratings. More specifically, for any $2$ raters (out of $3$ raters) rating the same example, we compute the absolute distance between the values they chose from the likert scale\footnote{In our interface the raters agree with a statement (see \cref{tab:eval_statements}) on a 5-point scale between \emph{disagree completely} to \emph{neither} to \emph{agree completely} which we map to $1\ldots5$. See \cref{fig:interface_cow_rated} for a screenshot.} and report the average for each dimension in \cref{tab:agreement-pair}. In general, we observe a reasonably high agreement among the raters, as the average distance is close to or below $1$ in most dimensions. Notably, we observe a higher agreement in the presentational dimensions \emph{style}, \emph{clarity}, and \emph{correctness}.

\begin{table}[htp]
\centering
\footnotesize
\begin{tabular}{lSSSSSSSSSS}
\toprule
Issue & \multicolumn{3}{c}{InstructGPT}  & \multicolumn{1}{c}{ChatGPT} & \multicolumn{1}{c}{PaLM-2} & \multicolumn{1}{c}{GPT4} & \multicolumn{1}{c}{Falcon}  \\
& \multicolumn{1}{c}{\footnotesize davinci-002} & \multicolumn{1}{c}{\footnotesize davinci-003} & \multicolumn{1}{c}{\footnotesize turbo} & \multicolumn{1}{c}{} & \multicolumn{1}{c}{\footnotesize text-bison} & \multicolumn{1}{c}{} & \multicolumn{1}{c}{\footnotesize 180B-Chat}  \\
\midrule
style                & 1.12 & 0.95 & 0.76 & 0.61 & 0.88 & 0.79 & 0.75 \\
clarity              & 0.97 & 0.74 & 0.73 & 0.59 & 0.69 & 0.81 & 0.69 \\
correctness          & 0.98 & 0.69 & 0.66 & 0.56 & 0.59 & 0.62 & 0.68 \\
tone                 & 1.16 & 1.26 & 1.21 & 1.30 & 1.36 & 1.22 & 1.23 \\
accuracy             & 1.05 & 0.97 & 1.07 & 1.15 & 1.13 & 0.97 & 0.95 \\
specificity          & 1.04 & 1.16 & 1.06 & 0.98 & 1.23 & 1.26 & 1.20 \\
completeness         & 1.00 & 1.03 & 1.06 & 0.71 & 1.13 & 1.01 & 1.21 \\
uncertainty          & 0.95 & 0.98 & 0.89 & 0.57 & 1.10 & 0.78 & 1.26 \\
\bottomrule
\end{tabular}
\caption{Average pairwise distance between likert ratings for each dimension. Distances between ratings on presentational adequacy are generally lower compared to epistemological adequacy.}
\label{tab:agreement-pair}
\end{table}

\textbf{Krippendorff's alpha.} In addition to pairwise distances, we compute Krippendorff's alpha. Krippendorff's alpha measures $1-\frac{D_\text{o}}{D_\text{e}}$, where $D_{\text{o}}$ is the observed disagreement, and $D_{\text{e}}$ is the expected disagreement by chance. Values are in $[-1, 1]$ range, where $1$ means complete agreement and $-1$ means complete systematic disagreement. Numbers in \cref{tab:agreement-alpha} suggest a similar trend to pairwise distance, where in most dimensions the agreement is medium, and the agreement in most presentational dimensions is higher compared to epistemological dimensions. 

\begin{table}[htp]
\centering
\footnotesize
\begin{tabular}{lSSSSSSSSSS}
\toprule
Issue & \multicolumn{3}{c}{InstructGPT}  & \multicolumn{1}{c}{ChatGPT} & \multicolumn{1}{c}{PaLM-2} & \multicolumn{1}{c}{GPT4} & \multicolumn{1}{c}{Falcon}  \\
& \multicolumn{1}{c}{\footnotesize davinci-002} & \multicolumn{1}{c}{\footnotesize davinci-003} & \multicolumn{1}{c}{\footnotesize turbo} & \multicolumn{1}{c}{} & \multicolumn{1}{c}{\footnotesize text-bison} & \multicolumn{1}{c}{} & \multicolumn{1}{c}{\footnotesize 180B-Chat}  \\
\midrule
style                & 0.45 & 0.53 & 0.74 & 0.70 & 0.60 & 0.48 & 0.72 \\
clarity              & 0.59 & 0.73 & 0.60 & 0.72 & 0.72 & 0.65 & 0.77 \\
correctness          & 0.57 & 0.74 & 0.80 & 0.85 & 0.82 & 0.71 & 0.78 \\
tone                 & 0.48 & 0.36 & 0.41 & 0.31 & 0.25 & 0.36 & 0.41 \\
accuracy             & 0.56 & 0.57 & 0.52 & 0.46 & 0.46 & 0.59 & 0.62 \\
specificity          & 0.53 & 0.40 & 0.50 & 0.51 & 0.32 & 0.32 & 0.39 \\
completeness         & 0.57 & 0.48 & 0.47 & 0.64 & 0.38 & 0.46 & 0.37 \\
uncertainty          & 0.59 & 0.51 & 0.57 & 0.75 & 0.40 & 0.63 & 0.32 \\
\bottomrule
\end{tabular}
\caption{Krippendorff's alpha of $3$ likert ratings per dimension. In general we observe a medium agreement. For most LLMs the value is higher for the presentational dimensions, except tone.}
\label{tab:agreement-alpha}
\end{table}

Note that either measure of agreement is subject to interpretability
shortcomings: Krippendorff's alpha can be misleadingly low in the case of low
overall variability, i.e.\ when many examples are rated as $5$ in a certain dimension. Likewise, average pairwise distance would appear too high.

Furthermore, we measure the agreement among raters when choosing issues. A rater might select or not select a given issue for a given answer, therefore, the value of interest is a binary variable. As above report two metrics of agreement:

\textbf{Pairwise agreement.} We look at the agreement among raters when selecting or not selecting a given issue. Particularly, we consider $2$ raters to agree with each other on a certain issue for a given answer if they both select or both not select that issue. We then report the percentage of pairwise agreement per issue in \cref{tab:agreement-pair-issue}. For the majority of issues we observe a high agreement among raters. As one might expect, issues such as ``not enough detail'', ``vague'', ``uncertainty missing'', and ``biased'' are more controversial and we see a lower agreement among the raters. 

\begin{table}
\centering
\footnotesize
\resizebox{\textwidth}{!}{%
    \begin{tabular}{lSSSSSSSSSS}
\toprule
Issue & \multicolumn{3}{c}{InstructGPT}  & \multicolumn{1}{c}{ChatGPT} & \multicolumn{1}{c}{PaLM-2} & \multicolumn{1}{c}{GPT4} & \multicolumn{1}{c}{Falcon}  \\
& \multicolumn{1}{c}{\scriptsize davinci-002} & \multicolumn{1}{c}{\scriptsize davinci-003} & \multicolumn{1}{c}{\scriptsize turbo} & \multicolumn{1}{c}{} & \multicolumn{1}{c}{\scriptsize text-bison} & \multicolumn{1}{c}{} & \multicolumn{1}{c}{\scriptsize 180B-Chat}  \\
\midrule
\textbf{style} \\
too informal                   & 92.40 & 97.77 & 99.55 & 99.78 & 96.42 & 97.10 & 98.21 \\
too long                       & 98.20 & 98.00 & 99.33 & 98.44 & 96.42 & 95.77 & 98.44 \\
too short                      & 84.79 & 87.42 & 99.66 & 99.55 & 95.08 & 99.33 & 97.77 \\
inconsistent                   & 90.72 & 98.00 & 99.33 & 100.00 & 97.76 & 99.55 & 99.11 \\
repetitive                     & 83.63 & 96.88 & 99.78 & 98.88 & 97.76 & 97.77 & 99.33 \\
\vspace{5pt} other             & 94.33 & 98.00 & 99.55 & 100.00 & 98.88 & 98.66 & 99.33 \\

\textbf{clarity} \\
sentences too long             & 96.39 & 97.77 & 99.78 & 99.55 & 96.64 & 94.21 & 97.54 \\
too technical                  & 94.07 & 98.22 & 99.33 & 99.11 & 97.76 & 95.10 & 98.88 \\
hard math                      & 96.91 & 99.11 & 97.31 & 99.33 & 98.66 & 96.88 & 100.00 \\
\vspace{5pt} other             & 85.95 & 98.22 & 99.55 & 99.78 & 97.76 & 99.11 & 98.44 \\

\textbf{correctness} \\
incomplete sentence            & 94.33 & 97.11 & 99.55 & 100.00 & 100.00 & 100.00 & 100.00 \\
incorrect spelling             & 98.45 & 100.00 & 99.78 & 99.78 & 99.55 & 100.00 & 99.78 \\
incorrect punctuation          & 95.88 & 99.11 & 98.88 & 100.00 & 99.78 & 99.55 & 98.88 \\
incorrect grammar              & 89.43 & 99.33 & 99.33 & 99.78 & 99.78 & 99.78 & 98.66 \\
\vspace{5pt} other             & 93.81 & 98.22 & 100.00 & 99.78 & 99.33 & 99.78 & 98.66 \\

\textbf{tone} \\
biased                         & 60.57 & 59.02 & 67.15 & 48.33 & 57.06 & 59.19 & 64.06 \\
persuasive                     & 95.62 & 87.08 & 91.03 & 86.38 & 81.05 & 84.98 & 91.96 \\
negative                       & 97.68 & 96.66 & 97.09 & 98.21 & 90.92 & 95.07 & 96.65 \\
\vspace{5pt} other             & 99.23 & 98.66 & 99.55 & 99.78 & 99.33 & 96.08 & 99.55 \\

\textbf{accuracy} \\
incorrect                      & 69.91 & 89.73 & 91.43 & 95.70 & 84.36 & 92.20 & 93.02 \\
science out of context         & 82.70 & 89.35 & 88.76 & 95.22 & 89.17 & 92.46 & 95.56 \\
self contradictory             & 95.49 & 98.20 & 99.87 & 99.76 & 98.40 & 98.98 & 99.24 \\
anecdotal                      & 78.05 & 97.18 & 92.37 & 63.80 & 87.17 & 94.63 & 94.67 \\
wrong use of terms             & 96.22 & 97.69 & 98.26 & 99.52 & 97.46 & 98.98 & 99.49 \\
\vspace{5pt} other             & 93.46 & 93.84 & 96.12 & 96.42 & 95.45 & 89.13 & 98.48 \\

\textbf{specificity} \\
irrelevant info                & 75.59 & 90.79 & 92.43 & 89.84 & 84.89 & 84.06 & 93.60 \\
vague                          & 48.96 & 54.20 & 60.44 & 58.90 & 52.60 & 56.81 & 60.66 \\
\vspace{5pt} other             & 97.13 & 93.94 & 97.71 & 99.09 & 96.31 & 97.00 & 97.27 \\

\textbf{completeness} \\
does not address main parts    & 61.33 & 69.27 & 80.68 & 82.77 & 75.12 & 84.07 & 84.95 \\
does not address region        & 93.36 & 94.78 & 98.41 & 98.87 & 97.30 & 96.59 & 97.92 \\
does not address time          & 96.09 & 91.61 & 96.14 & 98.64 & 98.59 & 94.54 & 98.73 \\
not enough detail              & 44.66 & 55.56 & 51.48 & 68.59 & 54.23 & 59.39 & 47.57 \\
ignores science                & 84.24 & 77.55 & 86.93 & 90.82 & 81.57 & 90.22 & 94.91 \\
\vspace{5pt} other             & 97.66 & 98.30 & 99.09 & 99.77 & 98.12 & 94.77 & 97.80 \\

\textbf{uncertainty} \\
uncertainty missing            & 49.35 & 63.46 & 63.07 & 80.02 & 60.10 & 65.39 & 50.57 \\
consensus missing              & 70.26 & 75.45 & 66.97 & 81.96 & 77.25 & 81.60 & 82.53 \\
contradicting evidence missing & 92.47 & 88.46 & 94.84 & 95.89 & 91.97 & 92.94 & 95.17 \\
other                          & 98.18 & 97.74 & 99.20 & 99.77 & 99.03 & 96.06 & 99.31 \\
\bottomrule
\end{tabular}%
}
\caption{Pairwise agreement among the $3$ raters per issue. In general we observe high agreement among raters in selecting issues for all models, while some issues such as ``vague'', ``biased'', ``not enough detail'', and ``uncertainty missing'' are more disagreed upon.}
    \label{tab:agreement-pair-issue}
\end{table}

\textbf{Krippendorff's alpha.} Similarly, we compute the Krippendorff's alpha for agreement on issues and observe a similar trend in \cref{tab:agreement-alpha-issue}.

\begin{table}[htbp]
\centering
\footnotesize
\resizebox{\textwidth}{!}{%
    \begin{tabular}{lSSSSSSSSSS}
\toprule
Issue & \multicolumn{3}{c}{InstructGPT}  & \multicolumn{1}{c}{ChatGPT} & \multicolumn{1}{c}{PaLM-2} & \multicolumn{1}{c}{GPT4} & \multicolumn{1}{c}{Falcon}  \\
& \multicolumn{1}{c}{\scriptsize davinci-002} & \multicolumn{1}{c}{\scriptsize davinci-003} & \multicolumn{1}{c}{\scriptsize turbo} & \multicolumn{1}{c}{} & \multicolumn{1}{c}{\scriptsize text-bison} & \multicolumn{1}{c}{} & \multicolumn{1}{c}{\scriptsize 180B-Chat}  \\
\midrule

\textbf{style} \\
too informal                   & 0.85 & 0.96 & 0.99 & 1.00 & 0.93 & 0.94 & 0.96 \\
too long                       & 0.96 & 0.96 & 0.99 & 0.97 & 0.93 & 0.92 & 0.97 \\
too short                      & 0.70 & 0.75 & 0.99 & 0.99 & 0.90 & 0.99 & 0.96 \\
inconsistent                   & 0.81 & 0.96 & 0.99 & 1.00 & 0.96 & 0.99 & 0.98 \\
repetitive                     & 0.67 & 0.94 & 1.00 & 0.98 & 0.96 & 0.96 & 0.99 \\
\vspace{5pt} other             & 0.89 & 0.96 & 0.99 & 1.00 & 0.98 & 0.97 & 0.99 \\

\textbf{clarity} \\
sentences too long             & 0.93 & 0.96 & 1.00 & 0.99 & 0.93 & 0.88 & 0.95 \\
too technical                  & 0.88 & 0.96 & 0.99 & 0.98 & 0.96 & 0.90 & 0.98 \\
hard math                      & 0.94 & 0.98 & 0.95 & 0.99 & 0.97 & 0.94 & 1.00 \\
\vspace{5pt} other             & 0.72 & 0.96 & 0.99 & 1.00 & 0.96 & 0.98 & 0.97 \\

\textbf{correctness} \\
incomplete sentence            & 0.89 & 0.94 & 0.99 & 1.00 & 1.00 & 1.00 & 1.00 \\
incorrect spelling             & 0.97 & 1.00 & 1.00 & 1.00 & 0.99 & 1.00 & 1.00 \\
incorrect punctuation          & 0.92 & 0.98 & 0.98 & 1.00 & 1.00 & 0.99 & 0.98 \\
incorrect grammar              & 0.79 & 0.99 & 0.99 & 1.00 & 1.00 & 1.00 & 0.97 \\
\vspace{5pt} other             & 0.88 & 0.96 & 1.00 & 1.00 & 0.99 & 1.00 & 0.97 \\

\textbf{tone} \\
biased                         & 0.21 & 0.18 & 0.34 & -0.03 & 0.14 & 0.18 & 0.28 \\
persuasive                     & 0.91 & 0.74 & 0.82 & 0.73 & 0.62 & 0.70 & 0.84 \\
negative                       & 0.95 & 0.93 & 0.94 & 0.96 & 0.82 & 0.90 & 0.93 \\
\vspace{5pt} other             & 0.98 & 0.97 & 0.99 & 1.00 & 0.99 & 0.92 & 0.99 \\

\textbf{accuracy} \\
incorrect                      & 0.40 & 0.79 & 0.82 & 0.91 & 0.67 & 0.85 & 0.85 \\
science out of context         & 0.64 & 0.78 & 0.77 & 0.90 & 0.78 & 0.84 & 0.90 \\
self contradictory             & 0.91 & 0.97 & 1.00 & 1.00 & 0.97 & 0.98 & 0.99 \\
anecdotal                      & 0.57 & 0.94 & 0.85 & 0.28 & 0.75 & 0.89 & 0.89 \\
wrong use of terms             & 0.93 & 0.95 & 0.97 & 0.99 & 0.94 & 0.98 & 0.99 \\
\vspace{5pt} other             & 0.86 & 0.88 & 0.92 & 0.93 & 0.91 & 0.78 & 0.97 \\

\textbf{specificity} \\
irrelevant info                & 0.51 & 0.81 & 0.85 & 0.80 & 0.70 & 0.67 & 0.87 \\
vague                          & -0.02 & 0.08 & 0.21 & 0.18 & 0.05 & 0.14 & 0.21 \\
\vspace{5pt} other             & 0.94 & 0.88 & 0.95 & 0.98 & 0.93 & 0.94 & 0.94 \\

\textbf{completeness} \\
does not address main parts    & 0.23 & 0.38 & 0.61 & 0.65 & 0.51 & 0.68 & 0.70 \\
does not address region        & 0.87 & 0.90 & 0.97 & 0.98 & 0.95 & 0.93 & 0.95 \\
does not address time          & 0.92 & 0.83 & 0.92 & 0.97 & 0.97 & 0.89 & 0.97 \\
not enough detail              & -0.11 & 0.11 & 0.03 & 0.38 & 0.09 & 0.19 & -0.05 \\
ignores science                & 0.68 & 0.55 & 0.73 & 0.82 & 0.63 & 0.80 & 0.90 \\
\vspace{5pt} other             & 0.95 & 0.96 & 0.98 & 1.00 & 0.96 & 0.90 & 0.95 \\

\textbf{uncertainty} \\
uncertainty missing            & -0.01 & 0.27 & 0.26 & 0.60 & 0.20 & 0.31 & 0.01 \\
consensus missing              & 0.41 & 0.51 & 0.33 & 0.64 & 0.53 & 0.64 & 0.65 \\
contradicting evidence missing & 0.85 & 0.77 & 0.90 & 0.92 & 0.84 & 0.86 & 0.91 \\
other                          & 0.96 & 0.96 & 0.98 & 1.00 & 0.98 & 0.92 & 0.99 \\
\bottomrule
\end{tabular}%
}
\caption{Krippendorff's alpha for agreement on issue selection. The results are consistent with patterns observed in pairwise agreement.}
    \label{tab:agreement-alpha-issue}
\end{table}

Looking at \cref{tab:issue_frequencies} we note that some issues are rarely chosen by raters and thus pairwise agreement numbers might be artificially high. For a deeper understanding regarding how well raters are able to agree on a specific issue we compute Krippendorff's alpha only for low ratings, i.e.\ cases where raters are required to select one or more issues. We report these numbers for a subset of dimensions with higher incidence counts in \cref{tab:agreement-alpha-issue-subset}. As hinted by incidence prevalence in  \cref{tab:issue_frequencies}, we find that when raters agree on a low rating for an epistemological dimension, they also exhibit medium to high agreement on what the specific issue is. One exception is \emph{accuracy:incorrect} which might be too generic as an issue.

Overall, agreement on specific issues is not high enough to recommend our 3-rater setup for evaluation of individual answers but for comparing and highlighting the strengths and shortcomings of models on a system level, as indicated by the fairly tight error bars in \cref{fig:all_results}.

\begin{table}[htbp]
\centering
\footnotesize
\resizebox{\textwidth}{!}{%
    \begin{tabular}{lSSSSSSSSSS}
\toprule
Issue & \multicolumn{3}{c}{InstructGPT}  & \multicolumn{1}{c}{ChatGPT} & \multicolumn{1}{c}{PaLM-2} & \multicolumn{1}{c}{GPT4} & \multicolumn{1}{c}{Falcon}  \\
& \multicolumn{1}{c}{\scriptsize davinci-002} & \multicolumn{1}{c}{\scriptsize davinci-003} & \multicolumn{1}{c}{\scriptsize turbo} & \multicolumn{1}{c}{} & \multicolumn{1}{c}{\scriptsize text-bison} & \multicolumn{1}{c}{} & \multicolumn{1}{c}{\scriptsize 180B-Chat}  \\
\midrule
\textbf{tone} \\
\vspace{5pt} biased                         & 0.84 & 0.60 & 0.55 & 0.58 & 0.34 & 0.35 & 0.43 \\

\textbf{accuracy} \\
\vspace{5pt} incorrect                      & 0.01 & 0.35 & 0.10 & 0.45 & 0.21 & 0.44 & 0.32 \\

\textbf{specificity} \\
\vspace{5pt} vague                          & 0.37 & 0.57 & 0.83 & 0.76 & 0.66 & 0.65 & 0.65 \\

\textbf{completeness} \\
does not address main parts    & 0.16 & 0.31 & 0.49 & 0.59 & 0.42 & 0.64 & 0.51 \\
\vspace{5pt} not enough detail              & 0.30 & 0.62 & 0.77 & 0.87 & 0.63 & 0.72 & 0.71 \\

\textbf{uncertainty} \\
uncertainty missing            & 0.55 & 0.81 & 0.77 & 0.91 & 0.83 & 0.80 & 0.72 \\
consensus missing              & 0.26 & 0.42 & 0.25 & 0.61 & 0.45 & 0.56 & 0.54 \\
\bottomrule
\end{tabular}%
}
\caption{Krippendorff's alpha for agreement on issue selection, but computed only for low ratings.}
    \label{tab:agreement-alpha-issue-subset}
\end{table}

\newpage
\subsection{AIS Evaluation} \label{sec:ais}
In our AIS evaluation we ask human raters to assess whether each keypoint is, fully, partially, or not supported by its corresponding top 3 retrieved passages. At the keypoint level, as shown in \cref{fig:ais}, the majority of keypoints are partially supported, while only $12\%$ of keypoints are fully supported by their corresponding evidence. We further ask the raters whether the top 3 passages jointly provide support for the keypoint. We then aggregate this result in the following way. We consider an answer to be fully attributed if all its keypoints are supported. An answer is not supported if all its keypoints are not supported. Otherwise, the answer is partially supported. An example is given in \cref{tab:ais-example}. We observe (\cref{fig:ais}) that only $6.95\%$ of answers are fully supported.
\begin{table}
\centering
\begin{tabular}{lcccc}\toprule
& \multicolumn{2}{c}{Per Example} & \multicolumn{2}{c}{Per Keypoint} \\ \cmidrule(l){2-3} \cmidrule(l){4-5}
& Percentage \% & Count & Percentage \% & Count  \\ \midrule
Fully Supports      & $6.95$ & $16$ & $12$ & $124$ \\
Partially Supports & $39.13$ & $90$ & $54.79$ & $566$ \\
No Support      & $53.91$ & $124$ & $32.81$ & $339$\\ 
Contradicts & $0$ & $0$ & $0.38$ & $4$ \\ \midrule
Total & $100$ & $230$ & $100$ & $1033$ \\
\bottomrule
\end{tabular}
\caption{AIS results for \gptfour{} generations. We report the results both at example level, where an example can consist of multiple keypoints, and at keypoint level. We observe that the majority of keypoints are only partially supported, and only around $7\%$ of examples are fully supported.}\label{tab:ais}
\end{table}
\begin{table}[htp]
\centering
\small
\begin{tabular}{@{}p{2cm}p{11cm}@{}}
\toprule
\textbf{Question} & How could pests and diseases spread more easily under warmer conditions caused by climate change? \\
\midrule
\textbf{Answer} & Climate change, particularly global warming, can facilitate the spread of pests and diseases more easily. \ctext{clipink}{Warmer conditions can extend the breeding seasons of pests, increase their population growth, and expand their geographical range.} For instance, \ctext{clip}{mosquitoes, which are vectors for diseases like malaria and dengue, thrive in warmer climates and can reproduce and mature more quickly under such conditions.} Additionally, \ctext{clip}{warmer temperatures can also accelerate the lifecycle of pathogens, making diseases more virulent and their transmission more efficient.} \\
\midrule
\ctext{clipink}{\textbf{1$^\text{st}$ Keypoint}} &  Warmer conditions can extend the breeding seasons of pests, increase their population growth, and expand their geographical range.\\
\midrule
\textbf{Paragraph 1.} & Infectious diseases that are sensitive to climate can be grouped into: vector-borne diseases (transmitted via mosquitos, ticks etc.), water-borne diseases (transmitted via viruses or bacteria, e.g.: E. Coli) and food-borne diseases (e.g.: Salmonella). Climate change is affecting the distribution of these diseases due to the expanding geographic range and seasonality of these diseases and their vectors. \\ \midrule
\textbf{Paragraph 2.} & Mosquito-borne diseases that are sensitive to climate include malaria, elephantiasis, Rift Valley fever, yellow fever, dengue fever, Zika virus, and chikungunya. Scientists found in 2022 that rising temperatures are increasing the areas where dengue fever, malaria and other mosquito-carried diseases are able to spread. Warmer temperatures are also advancing to higher elevations, allowing mosquitoes to survive in places that were previously inhospitable to them. This risks malaria  making a return to areas where it was previously eradicated. \\ \midrule
\textbf{Paragraph 3.} & Ticks are changing their geographic range because of rising temperatures, and this puts new populations at risk. Ticks can spread lyme disease and tick-borne encephalitis. It is expected that climate change will increase the incidence of these diseases in the Northern Hemisphere. For example, a review of the literature found that "In the USA, a 2°C warming could increase the number of Lyme disease cases by over 20\% over the coming decades and lead to an earlier onset and longer length of the annual Lyme disease season". \\ \midrule
\textbf{Rating} & Not Supported \\ \midrule
\ctext{clip}{\textbf{2$^\text{nd}$ Keypoint}} & Mosquitoes, which are vectors for diseases like malaria and dengue, thrive in warmer climates and can reproduce and mature more quickly under such conditions. \\
\textbf{Paragraphs} & $\dots$ \\
\textbf{Rating} & Supported \\ \midrule
\ctext{clip}{\textbf{3$^\text{rd}$ Keypoint}} & Warmer temperatures can also accelerate the lifecycle of pathogens, making diseases more virulent and their transmission more efficient.\\
\textbf{Paragraphs} & $\dots$ \\
\textbf{Rating} & Supported \\ \midrule
\textbf{Overall Rating} & Partially Supported \\
\bottomrule
\end{tabular}
\caption{An example of the AIS task with ratings. Overall, the answer is only partially supported, as the first keypoint is not supported by the evidence.}
\label{tab:ais-example}
\end{table}
\begin{table}
\footnotesize
\centering
\begin{tabular}{p{6cm}p{6cm}}\toprule
\multicolumn{2}{p{12cm}}{\textbf{Question.} When did the general public become aware of carbon dioxide's effects on climate?}  \\ \midrule
\textbf{Answer.} The general public became significantly aware of carbon dioxide's effects on climate change around the late 20th century. The Intergovernmental Panel on Climate Change (IPCC) was established in 1988, and its assessments, which highlighted the role of carbon dioxide in global warming, were widely disseminated and discussed in the media. However, it's important to note that the scientific understanding of carbon dioxide's impact on climate predates this public awareness by many decades, with initial theories and evidence emerging as early as the late 19th century. & \textbf{AI Assistance.} While the answer provides a general timeline of public awareness of carbon dioxide's effects on climate change, it fails to mention key milestones such as the warnings of the climate crisis by Glenn T. Seaborg in 1966, or John Sawyer's study in 1972, which summarized the knowledge of the science at the time and accurately predicted the rate of global warming for the period between 1972 and 2000. These events significantly contributed to the public's understanding of the issue before the establishment of the IPCC in 1988. \\ 
\textbf{AIS.} \colorbox{clip}{Fully attributable} & \textbf{Average completeness rating.} \colorbox{clipink}{1.6} \\ \midrule
\multicolumn{2}{p{12cm}}{\textbf{Question.} How do elevated carbon dioxide levels affect the mineral density of crops and wild plants?}  \\\midrule
\textbf{Answer.} Elevated carbon dioxide levels can significantly affect the mineral density of crops and wild plants. While higher CO2 levels can stimulate plant growth, they can also reduce the concentration of essential minerals in a variety of crops and wild plants. This phenomenon, known as "CO2 dilution effect", results in crops with lower nutritional value, including reduced levels of protein, iron, zinc, and other essential nutrients. Therefore, while plants may grow larger in response to increased CO2, their nutritional quality can be significantly compromised. & \textbf{AI Assistance.} The statement fails to acknowledge the uncertainty in the potential positive impacts of elevated CO2 levels, such as higher crop yields due to CO2 fertilization, and the possibility of more arable land becoming available as frozen land thaws. \\ 
\textbf{AIS.} \colorbox{clip}{Fully attributable} & \textbf{Average uncertainty rating.} \colorbox{clipink}{1.6} \\ \midrule
\multicolumn{2}{p{12cm}}{\textbf{Question.} Can climate sensitivity be ignored because too low?}  \\ \midrule
\multicolumn{2}{p{12cm}}{\textbf{Answer.} Climate sensitivity, which refers to the temperature change associated with a doubling of the concentration of carbon dioxide in the atmosphere, cannot be ignored even if it is low. This is because even a small increase in global temperatures can have significant impacts on the Earth's climate system. For instance, it can lead to more frequent and severe weather events, such as hurricanes and droughts, and cause sea levels to rise, which can result in coastal flooding. Therefore, understanding and addressing climate sensitivity is crucial in climate change mitigation and adaptation strategies.} \\
\textbf{AIS.} \colorbox{clipink}{Not Supported} & \textbf{Average accuracy rating.} \colorbox{clip}{4.7} \\ \midrule
\multicolumn{2}{p{12cm}}{\textbf{Question.} Is global warming caused by the pacific decadal oscillation?}  \\ \midrule
\multicolumn{2}{p{12cm}}{\textbf{Answer.} No, global warming is not caused by the Pacific Decadal Oscillation (PDO). The PDO is a long-term ocean temperature fluctuation in the Pacific Ocean that shifts between warm and cool phases approximately every 20 to 30 years. While these shifts can influence regional climate patterns and marine ecosystems, they do not cause global warming. Global warming is primarily caused by human activities, particularly the burning of fossil fuels like coal, oil, and gas, which increase the concentration of greenhouse gases in the atmosphere and trap heat.} \\
\textbf{AIS.} \colorbox{clipink}{Not Supported} & \textbf{Average specificity rating.} \colorbox{clip}{5} \\
\bottomrule
\end{tabular}
\caption{Examples highlighting the differences between attribution scores and epistemological ratings under our framework.}
\label{tab:ais-examples}
\end{table}

\newpage
\subsection{Breakdown of Ratings Per Question Type}
\label{sec:rating_by_type}
\begin{figure}[htbp]
    \centering
      \includesvg[width=\textwidth]{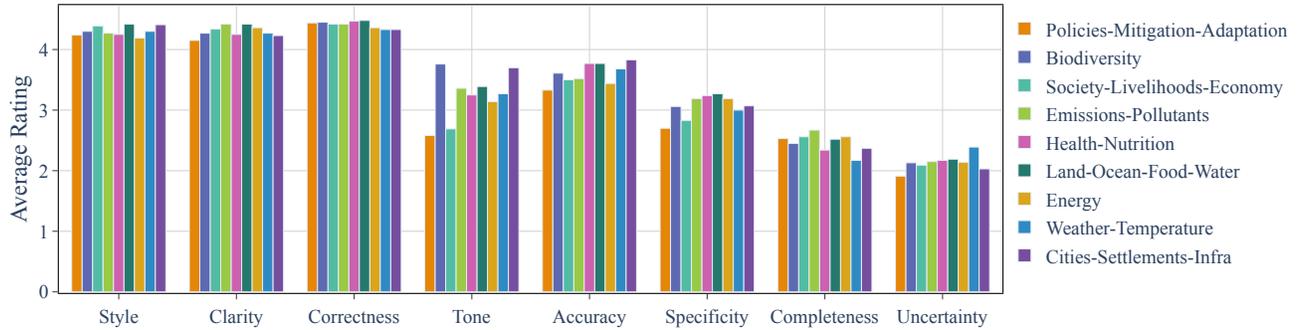}
    \caption{Average rating along all the dimensions per question topic. Questions in the \textit{``Policies-Mitigation-Adaption''} category receive lower ratings in most of the epistemological dimensions, and particularly in ``Tone'' dimension.}
    \label{fig:rating-per-type}
\end{figure}

\begin{figure}[htbp]
    \centering
      \includesvg[width=0.8\textwidth]{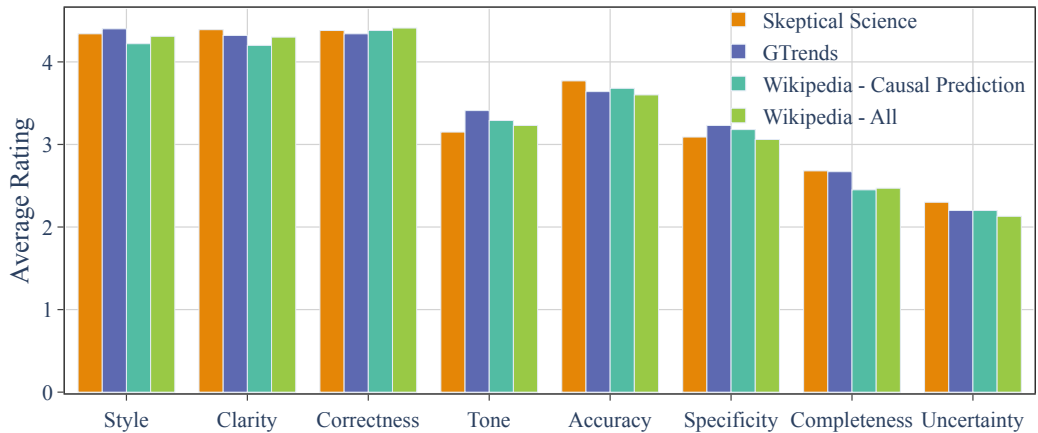}
    \caption{Average rating along all the dimensions per question source and type. In general, there is not a significant difference among the ratings based on the question source or causality.}
    \label{fig:rating-per-q}
\end{figure}

We compare the presentational and epistemological adequacy of \gptfour{} answers, based on the question source, type, and causal-prediction dimension, as described in \cref{sec:synthetic-questions}. Generally, there isn't a significant difference between the ratings based on the topic of the question as shown in \cref{fig:rating-per-type}. However, we observe that questions in the \textit{``Policies-Mitigation-Adaptation''} category receive lower ratings in most of the epistemological dimensions, and particularly in the \emph{tone} dimension.
We further look at the difference in average ratings based on the source of the question (Wikipedia, Skeptical Science, or GTrends), and causality of the question. The source of the question does not affect the ratings significantly (please refer to \cref{fig:rating-per-q}). However, we observe that Wikipedia questions tend to receive lower epistemological adequacy ratings. This could be because these questions ask for more details and very specific info compared to GTrends and Skeptical Sciences, and thus are harder to answer.

\newpage
\subsection{Validation experiments}
\label{sect:validation-details}

To validate our evaluation framework, we manually created a set of $30$ questions
and answers, each exhibiting a specific issue from our full list of issues. We pose this as a task to our raters in two settings: with and without AI Assistance, each replicated three times.

\begin{table}[t]
\centering
\begin{tabular}{lS@{\hskip 5pt}lS@{\hskip 5pt}l}\toprule
         &  \multicolumn{2}{c}{Without AI Assistance (3x)} &  \multicolumn{2}{c}{With AI assistance (3x)}  \\ 
\midrule
Any      & 83.33 & $\%$ & 93.33 & $\%$ \\
Majority & 60.00 & $\%$ & 76.67 & $\%$ \\
All      & 33.33 & $\%$ & 43.33 & $\%$ \\ 
\bottomrule
\end{tabular}
\caption{Validation results. The percentage of the (30) issues recognized by any rater, the majority of raters or all of the raters.}\label{tab:validation}
\end{table}

We found (cf. \cref{tab:validation}) that without assistance, most of the issues ($83\%$) were found by at least one rater, but only $60\%$ of issues by the majority of raters.

With the help of assistance, more issues were detected ($93.33\%$ were detected by at least one rater, $76.67\%$ by the majority of raters). There was just one issue not detected by any rater: \emph{anecdotal evidence}\footnote{All raters did correctly rate the dimension (`accuracy' as 1 or 2), but highlighted only the 'incorrect' issue. The answer was incorrect, based on anecdotal evidence.}.

\newpage
\subsection{LLM Rater}
\label{sect:appendix-llm-rater}

We investigate the possibility of using an LLM to perform the rating task on our evaluation framework. We use \gptfour{} and prompt it using the same language as presented to the human raters (\cref{tab:llm_rater}). We sample 3 responses (temperature 0.6) from \gptfour{} for each question to replicate the setup we have with human raters. We observe the following from the results in \cref{tab:main_results_llm_rater}. First, \gptfour{} rater also benefits from assistance. Consistent with findings with human raters, \gptfour{} rater rates answers lower when assistance is provided. This makes sense because (1) the assistance provides an additional chain-of-thought like input and (2) the assistance is generated using additional documents which potentially provide the model with additional information. When the issues in the answers are more severe, \gptfour{} rater agrees with human raters on all dimensions, as evident in the ratings for \gpttwo{}. However, \gptfour{} rater disagrees with human raters on the relative ranking of answers from different models of similar quality. Notably, the \gptfour{} rater is more generous towards OpenAI models than humans are.

\begin{table}[htbp]
\centering
\footnotesize
\begin{tabular}{@{}p{3.4cm}p{9.5cm}@{}}\toprule
         Function & Prompt  \\ \midrule
System & You are an expert on climate communication. Given a question, and an answer to it, and maybe a succinct critique of the answer. \newline \\
Rating & To what extent do you agree with the following statement on the 5-point likert scale (1=disagree completely, 2=disagree, 3=neither, 4=agree, 5=agree completely)?\newline Question: [\textit{question}]\newline Answer: [\textit{answer}]\newline Critique: [\textit{critique}]\newline Statement: [\textit{statement for different dimensions (below)}]\newline\newline Answer in the format "Rating: X Problem: Y Explanation: Z" \\
\midrule
Style Statement & The information is presented well (for a general audience).\newline If you disagree, what is the problem with the answer? Choose one of the following: too informal/colloquial, answer too long, answer too short, inconsistent language/style/terminology, repetitive, other.\newline If you choose other, please explain your rating.\newline \\
Clarity Statement & The answer is clear and easy to understand.\newline If you disagree, what is the problem with the answer? Choose one of the following: sentences too long, language too technical, numbers/formulae hard to understand, other.\newline If you choose other, please explain your rating.\newline \\
Correctness Statement & The language in the answer does not contain mistakes.\newline If you disagree, what is the problem with the answer? Choose one of the following: sentence is incomplete, spelling mistakes, punctuation mistakes, grammatical errors, other.\newline If you choose other, please explain your rating.\newline \\
Tone Statement & The tone of the answer is neutral and unbiased.\newline If you disagree, what is the problem with the answer? Choose one of the following: the answer is biased, tries to convince me of an opinion/belief, the tone is too negative, other.\newline If you choose other, please explain your rating. \\
\midrule
Accuracy Statement & The answer is accurate.\newline If you disagree, what is the problem with the answer? Choose one of the following: incorrect, takes scientific findings out of context, self-contradictory, anecdotal, wrong use of key terms/scientific terminology, other.\newline If you choose other, please explain your rating.\newline \\
Specificity Statement & The answer addresses only what the question asks for, without adding irrelevant information.\newline If you disagree, what is the problem with the answer? Choose one of the following: includes irrelevant parts, too vague/unspecific, other.\newline If you choose other, please explain your rating.\newline \\ 
Completeness Statement & The answer addresses everything the question asks for.\newline If you disagree, what is the problem with the answer? Choose one of the following: misses important parts of the answer, does not address the region the question asks about, does not address time or time range the question asks about, does not give enough detail (e.g., numbers, statistics, details), ignores relevant scientific knowledge, other.\newline If you choose other, please explain your rating.\newline \\
Uncertainty Statement & The answer appropriately conveys the uncertainty involved.\newline If you disagree, what is the problem with the answer? Choose one of the following: degree of (un)certainty not given when it should be, agreement in the scientific community not given when important, contradicting evidence (if existing) not mentioned, other.\newline If you choose other, please explain your rating. \\ 

\bottomrule
\end{tabular}
\caption{Prompts used to generate ratings.}\label{tab:llm_rater}
\end{table}

\begin{table}[htp]
\centering
\small
\begin{tabular}{lc@{\hskip 4pt}c@{\hskip 4pt}c@{\hskip 4pt}c@{\hskip 4pt}c@{\hskip 4pt}c@{\hskip 4pt}c@{\hskip 4pt}c}
\toprule
System & style & clarity & correctness & tone & accuracy & specificity & completeness & uncertainty \\
\midrule
GPT4 & 4.71 & 4.89 & 5.00 & 3.88 & 4.13 & 3.66 & 2.97 & 2.05 \\
ChatGPT & 4.75 & 4.91 & 4.99 & 3.91 & 4.18 & 3.68 & 2.72 & 2.00 \\
InstructGPT (davinci-003) & 4.39 & 4.68 & 4.63 & 4.05 & 3.49 & 3.29 & 2.44 & 1.91 \\
InstructGPT (davinci-002) & 2.88 & 3.25 & 3.54 & 3.11 & 2.32 & 2.27 & 1.89 & 1.74 \\
InstructGPT (turbo) & 4.62 & 4.82 & 4.89 & 3.80 & 3.76 & 3.30 & 2.46 & 1.94 \\
PaLM-2 (text-bison) & 4.40 & 4.72 & 4.75 & 3.42 & 3.38 & 3.03 & 2.31 & 1.92 \\
Falcon (180B-Chat) & 4.66 & 4.85 & 4.91 & 3.83 & 4.03 & 3.49 & 2.71 & 2.00 \\
\midrule
GPT4, no assistance & 4.70 & 4.89 & 5.00 & 4.77 & 4.95 & 4.59 & 4.59 & 2.63 \\
\bottomrule
\end{tabular}
    \caption{Results from the LLM Rater.}
    \label{tab:main_results_llm_rater}
\end{table}

\newpage
\subsection{Timing analysis}
\label{sec:timing}
\begin{figure}[htbp]
    \centering
    \includesvg[width=\textwidth]{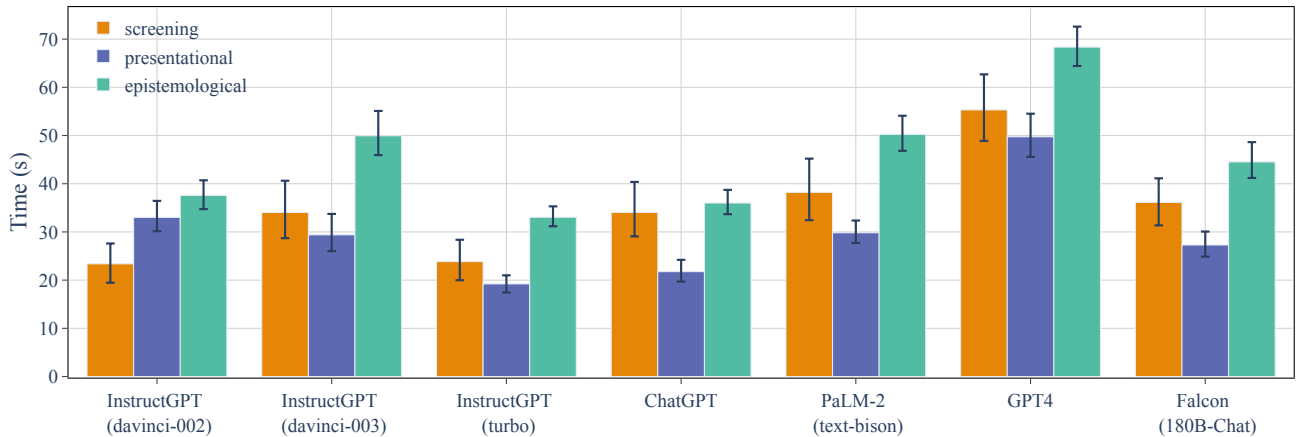} %
    \caption{Average time per example for the screening, presentational, and epistemological assessment.}
    \label{fig:time_per_dimtype}
\end{figure}

We analyze how long raters take for their tasks. As can be seen in \cref{fig:time_per_dimtype}, rating the epistemological dimensions generally takes more time than assessing the presentation quality, even though the latter is done first in our questionnaire. We also observe that for most systems the screening part, which includes the initial reading of question and answer, takes longer than rating the presentational dimensions. The exception to this rule are answers from \emph{InstructGPT (davinci-002)} which are often shorter and thus quicker to read.

\begin{figure}
    \centering
    \includesvg[width=.8\textwidth]{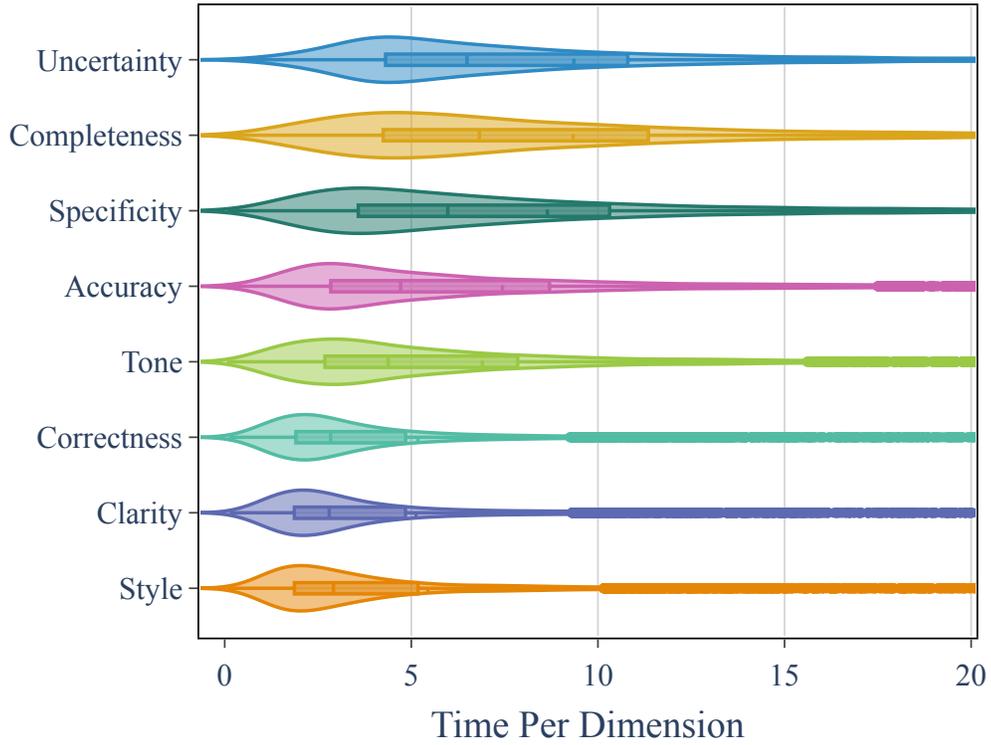} %
    \caption{Distribution of rating times for presentational (style, clarity, correctness, tone) and epistemological (specificity, uncertainty, completeness, accuracy) dimensions. For ease of presentation, this figure ignores a small number of timings that took longer than $60s$.}
    \label{fig:time_per_dim}
\end{figure}

\cref{fig:time_per_dim} shows that \emph{tone} seems to be harder to assess among the presentational dimensions whereas \emph{accuracy} is quicker among the epistemological dimensions. Otherwise, each dimension takes a similar amount of time.

\begin{figure}[htbp]
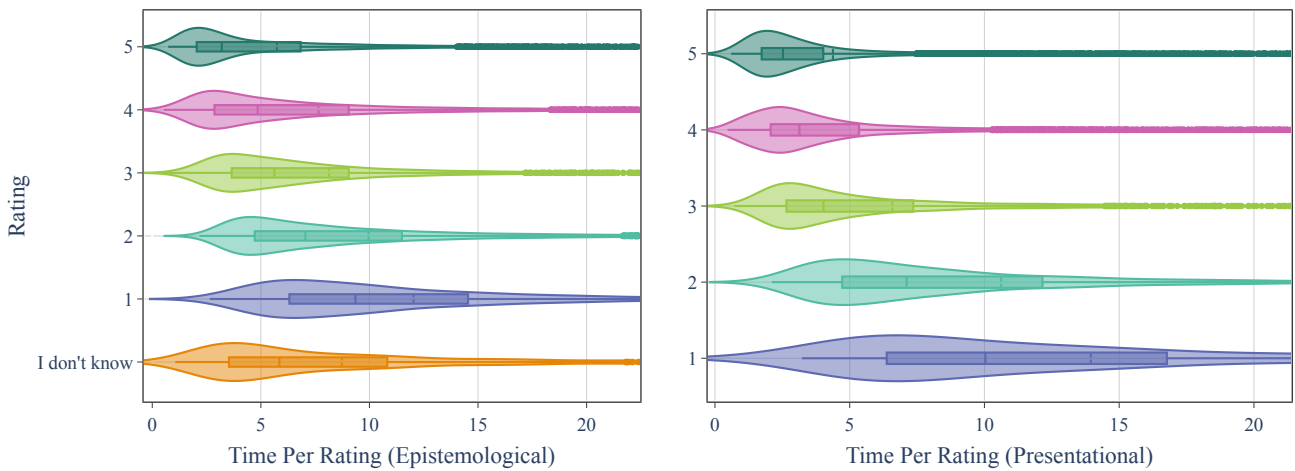

    \centering %
    \begin{subfigure}[b]{0.495\textwidth}
         \centering
         \includesvg[width=\textwidth]{figures/time-per-rating-epistemic.svg} 
     \end{subfigure}
     \hfill
     \begin{subfigure}[b]{0.495\textwidth}
         \centering
         \includesvg[width=\textwidth]{figures/time-per-rating-presentational.svg} 
     \end{subfigure}
    \caption{Average time per example for the epistemological and presentational assessment.}
    \label{fig:time_per_rating}
\end{figure}

Larger differences are revealed when we analyze how the rating itself affects the rating times. As expected, \cref{fig:time_per_rating} shows that high ratings are quicker than lower ones. Keep in mind that for disagreeing ratings (less than 3) we also require the raters to point out specific issues which may add to the length of the interaction. Nevertheless, the trend is also clear among the better (3-5) ratings as well as \emph{between} 1 and 2. For the epistemological dimensions the raters can also select \emph{I don't know}, which takes slightly longer than choosing the middle rating of 3.

\newpage

\subsection{Answers}
In this section, we present examples showing good versus bad answers for each dimension evaluated in Tables \ref{tab:example_presentation} to \ref{tab:example_uncertainty}.

\begin{table}[htp]
\centering
\small
\begin{tabular}{@{}p{1.2cm}p{6cm}p{6cm}@{}}
\toprule
\textbf{Question} & \multicolumn{2}{l}{In what ways can peri-urban ecosystems help reduce the urban heat island effect?} \\
\midrule
& \multicolumn{1}{c}{\textbf{Model A}} & \multicolumn{1}{c}{\textbf{Model B}} \\
\midrule
\textbf{Answer} & Peri-urban ecosystems can significantly help reduce the urban heat island effect in several ways. Firstly, they act as natural coolants by providing shade and releasing moisture into the atmosphere through a process known as evapotranspiration, which can lower surrounding temperatures. Secondly, these ecosystems, which include forests, grasslands, and wetlands, can absorb and store heat, reducing the overall temperature in urban areas. Lastly, they can also help mitigate the heat island effect by reducing the need for energy-intensive cooling systems, as well-vegetated areas are naturally cooler than concrete or asphalt surfaces. & Peri-urban ecosystems can help reduce the urban heat island effect by providing shade, evaporative cooling, and habitat for cooling breezes. Trees and other vegetation in peri-urban areas can help to block the sun's rays, reducing the amount of heat that is absorbed by the ground. The evaporation of water from plants and soil can also help to cool the air, and vegetation can create shade that helps to block the sun's heat. In addition, vegetation can provide habitat for animals that help to create cooling breezes. All of these factors can help to reduce the urban heat island effect, making cities more livable and sustainable. \\
\midrule
\textbf{Assistance} &  & The answer contains repetition, specifically in the explanation of how vegetation provides shade and blocks the sun's heat. \\
\midrule
\textbf{Issues} & & repetitive \\
\midrule
\textbf{Rating} & 4.6 & 2.0 \\
\bottomrule
\end{tabular}
\caption{Example for the \textbf{style} dimension. The Model B answer repeats the explanation on how vegetation provides shade and blocks the sun's heat.}
\label{tab:example_presentation}
\end{table}

\begin{table}[htp]
\centering
\small
\begin{tabular}{@{}p{1.2cm}p{6cm}p{6cm}@{}}
\toprule
\textbf{Question} & \multicolumn{2}{l}{Was the name 'global warming' changed to 'climate change'?} \\
\midrule
& \multicolumn{1}{c}{\textbf{Model A}} & \multicolumn{1}{c}{\textbf{Model B}} \\
\midrule
\textbf{Answer} & Yes & Yes, the term "global warming" was changed to "climate change" in order to better reflect the broader range of impacts associated with rising greenhouse gas emissions. While global warming refers specifically to the increase in Earth's average surface temperature, climate change encompasses a wider array of changes, including shifts in precipitation patterns, sea level rise, and the frequency and intensity of extreme weather events. The term "climate change" is now widely used to capture the full scope of these impacts. \\
\midrule
\textbf{Assistance} & The answer lacks sufficient detail to fully address the complexity of the question. &  \\
\midrule
\textbf{Issues} & too\_short, vague & \\
\midrule
\textbf{Rating} & 2.6 & 4.6 \\
\bottomrule
\end{tabular}
\caption{Example for the \textbf{clarity} dimension. The Model A answer is too short. It lacks details and context to fully address the question.}
\label{tab:example_clarity}
\end{table}

\begin{table}[htp]
\centering
\small
\begin{tabular}{@{}p{1.2cm}p{6cm}p{6cm}@{}}
\toprule
\textbf{Question} & \multicolumn{2}{l}{Which areas in Nigeria are at risk of rising sea levels?} \\
\midrule
& \multicolumn{1}{c}{\textbf{Model A}} & \multicolumn{1}{c}{\textbf{Model B}} \\
\midrule
\textbf{Answer} & The coastal areas of Nigeria are at risk of rising sea levels due to climate change. These areas include Lagos, Ogun, and Delta states. Rising sea levels could cause flooding, erosion, and salinization of freshwater sources. This could have a devastating impact on the livelihoods of the people who live in these areas. & I do not have access to the latest information on the specific areas in nigeria that are at risk of rising sea levels. however, it is important to note that nigeria, like many other countries, is vulnerable to the impacts of climate change, including sea level rise. coastal cities and communities are particularly at risk, and it is crucial for nigeria to take action to mitigate and adapt to the effects of climate change. \\
\midrule
\textbf{Assistance} &  & The answer contains a grammatical error - "nigeria" should be capitalized as "Nigeria". \\
\midrule
\textbf{Issues} & & incorrect\_grammar, incorrect\_spelling \\
\midrule
\textbf{Rating} & 4.6 & 2.3 \\
\bottomrule
\end{tabular}
\caption{Example for the \textbf{correctness} dimension. The Model B answer contains a grammatical error where the first letter of a location name should be capitalized.}
\label{tab:example_correctness}
\end{table}

\begin{table}[htp]
\centering
\small
\begin{tabular}{@{}p{1.2cm}p{6cm}p{6cm}@{}}
\toprule
\textbf{Question} & \multicolumn{2}{l}{Do volcanos emit more CO2 than humans?} \\
\midrule
& \multicolumn{1}{c}{\textbf{Model A}} & \multicolumn{1}{c}{\textbf{Model B}} \\
\midrule
\textbf{Answer} & No, volcanoes do not emit more CO2 than humans. According to the U.S. Geological Survey, human activities, particularly the burning of fossil fuels for energy, emit more than 60 times the amount of carbon dioxide than volcanoes each year. While volcanoes do contribute to CO2 emissions, their contribution is relatively small compared to human activities. It's important to note that human-induced climate change is primarily driven by the increase in greenhouse gases from human activities, not natural processes like volcanic eruptions. & No, volcanoes do not emit more CO2 than humans. While volcanic eruptions do release significant amounts of CO2 into the atmosphere, the total annual emissions from volcanoes are estimated to be about 0.3 to 0.4 billion metric tons. In contrast, human activities, such as burning fossil fuels and deforestation, emit over 35 billion metric tons of CO2 annually, which is about 100 times more than volcanic emissions. \\
\midrule
\textbf{Assistance} &  &  \\
\midrule
\textbf{Issues} & biased, persuasive &  \\
\midrule
\textbf{Rating} & 2.0 & 4.3 \\
\bottomrule
\end{tabular}
\caption{Example for the \textbf{tone} dimension. Instead of answering with only facts, the Model A answer assigns blame on human activities as the cause of climate change which was not part of the question.}
\label{tab:example_tone}
\end{table}

\begin{table}[htp]
\centering
\small
\begin{tabular}{@{}p{1.2cm}p{6cm}p{6cm}@{}}
\toprule
\textbf{Question} & \multicolumn{2}{l}{How does burning fossil fuels mitigate climate change} \\
\midrule
& \multicolumn{1}{c}{\textbf{Model A}} & \multicolumn{1}{c}{\textbf{Model B}} \\
\midrule
\textbf{Answer} & Burning fossil fuels does not mitigate climate change. In fact, it is one of the primary causes of climate change. When fossil fuels are burned, they release carbon dioxide and other greenhouse gases into the atmosphere, which trap heat and contribute to global warming. To mitigate climate change, we need to transition to cleaner, renewable sources of energy and reduce our overall energy consumption. & Burning fossil fuels releases greenhouse gases into the atmosphere, which trap heat and cause the planet to warm. This is the leading cause of climate change. However, burning fossil fuels can also be used to mitigate climate change by generating electricity from renewable sources, such as solar and wind power. This can help to reduce the amount of greenhouse gases emitted into the atmosphere and slow the rate of climate change. \\
\midrule
\textbf{AI Assistance} &  & The critique is that the statement inaccurately suggests that burning fossil fuels can mitigate climate change. As per the provided paragraphs, "the burning of fossil fuels is the main source of greenhouse gas emissions causing global warming and ocean acidification." Furthermore, "the International Energy Agency concluded that no new fossil fuel extraction projects could be opened if the global economy and society wants to avoid the worst impacts of climate change and meet international goals for climate change mitigation." Therefore, burning fossil fuels contributes to climate change rather than mitigating it.  \\
\midrule
\textbf{Issues} &  & incorrect, science\_out\_of\_context, self\_contradictory \\
\midrule
\textbf{Rating} & 4.6 & 1.3 \\
\bottomrule
\end{tabular}
\caption{Example for the \textbf{accuracy} dimension. The Model B answer contains a factual error which suggests that burning fossil fuels can mitigate climate change.}
\label{tab:example_accuracy}
\end{table}

\begin{table}[htp]
\centering
\small
\begin{tabular}{@{}p{1.2cm}p{6cm}p{6cm}@{}}
\toprule
\textbf{Question} & \multicolumn{2}{l}{What happened to global temperatures by the mid-1970s?} \\
\midrule
& \multicolumn{1}{c}{\textbf{Model A}} & \multicolumn{1}{c}{\textbf{Model B}} \\
\midrule
\textbf{Answer} & By the mid-1970s, global temperatures began to rise at an accelerated rate. This period marked the end of a slight cooling phase that had occurred from the 1940s to the 1970s, often referred to as the "global cooling" period. The shift in temperature trends was largely due to increased levels of greenhouse gases, particularly carbon dioxide, in the Earth's atmosphere from human activities such as burning fossil fuels. This marked the beginning of a period of rapid global warming that continues today. & By the mid-1970s, global temperatures had risen by about 0.2 degrees Celsius (0.36 degrees Fahrenheit) since the late 19th century. This warming trend was statistically significant, and it was consistent with the predictions of climate models that had been developed in the 1960s and 1970s. \\
\midrule
\textbf{Assistance} & The statement is incorrect in asserting that there are no vague or generic statements in the answer. The phrase "increased levels of greenhouse gases, particularly carbon dioxide, in the Earth's atmosphere from human activities such as burning fossil fuels" is a generic statement that could benefit from specific data or examples. As per Wikipedia, "The 2011–2020 decade warmed to an average 1.09 °C [0.95–1.20 °C] compared to the pre-industrial baseline (1850–1900). Surface temperatures are rising by about 0.2 °C per decade, with 2020 reaching a temperature of 1.2 °C above the pre-industrial era." This specific information could enhance the answer's clarity and precision. & The answer could be improved by including the specific information that "The 2011–2020 decade warmed to an average 1.09 °C [0.95–1.20 °C] compared to the pre-industrial baseline (1850–1900)" and "Surface temperatures are rising by about 0.2 °C per decade," which provides a more precise context for the warming trend observed by the mid-1970s.  \\
\midrule
\textbf{Issues} & vague &  \\
\midrule
\textbf{Rating} & 2 & 4 \\
\bottomrule
\end{tabular}
\caption{Example for the \textbf{specificity} dimension. The Model A answer omits specific numbers on how much the global temperatures have changed which is important for the answer's clarity and precision.}
\label{tab:example_specificity}
\end{table}

\begin{table}[htp]
\centering
\small
\begin{tabular}{@{}p{1.2cm}p{6cm}p{6cm}@{}}
\toprule
\textbf{Question} & \multicolumn{2}{l}{How much do cows contribute to climate change} \\
\midrule
& \multicolumn{1}{c}{\textbf{Model A}} & \multicolumn{1}{c}{\textbf{Model B}} \\
\midrule
\textbf{Answer} & Cows significantly contribute to climate change due to the production of methane, a potent greenhouse gas, during their digestion process. According to the United Nations' Food and Agriculture Organization, livestock (including cows) account for about 14.5\% of all human-caused greenhouse gas emissions, with cows being the largest contributor. This is equivalent to the exhaust emissions from all the vehicles on the planet. Therefore, the impact of cows on climate change is substantial and cannot be overlooked. & Cows are a major source of methane, a powerful greenhouse gas that contributes to climate change. According to the Environmental Protection Agency (EPA), cattle and other livestock are responsible for about 5.5 percent of all human-caused greenhouse gas emissions. \\
\midrule
\textbf{Assistance} & While the answer provides a comprehensive overview of cows' contribution to climate change, it fails to mention the potential for mitigation efforts, such as the adoption of best practices that can reduce emissions by up to 30\%, as stated by the FAO. Furthermore, it does not address the variability in environmental impacts due to different agricultural practices around the world. & While the answer provides a general overview of the contribution of cows to climate change, it fails to mention the potential mitigation strategies, such as dietary changes and adoption of best practices, which can significantly reduce emissions. As stated in the provided paragraphs, "Mitigation options for reducing methane emission from livestock include a change in diet, that is consuming less meat and dairy" and "FAO estimates that the adoption of already available best practices can reduce emissions by up to 30\%".  \\
\midrule
\textbf{Issues} &  & not\_enough\_detail, does\_not\_address\_main\_parts, ignores\_science \\
\midrule
\textbf{Rating} & 4.6 & 2 \\
\bottomrule
\end{tabular}
\caption{Example for the \textbf{completeness} dimension. The Model B answer lacks context and details around the answer. The assistance on missing mitigation efforts is not helpful because it is not necessary to make the answer complete. The raters are able to recognize this and give Model A answer a correct high rating.}
\label{tab:example_completeness}
\end{table}

\begin{table}[htp]
\centering
\small
\begin{tabular}{@{}p{1.2cm}p{6cm}p{6cm}@{}}
\toprule
\textbf{Question} & \multicolumn{2}{l}{Did Hansen predict that the West Side highway would be underwater in 20 years?} \\
\midrule
& \multicolumn{1}{c}{\textbf{Model A}} & \multicolumn{1}{c}{\textbf{Model B}} \\
\midrule
\textbf{Answer} & James Hansen, a prominent climate scientist, did make a statement in a 1988 interview suggesting that the West Side Highway in New York City could be underwater in 40 years. However, this was not a formal prediction based on a specific scientific study, but rather a speculative comment made to illustrate potential impacts of climate change. It's important to note that while sea levels are indeed rising due to climate change, the exact rate and impact can vary greatly depending on a range of factors, including greenhouse gas emissions and ice melt rates. & No, Hansen did not predict that the West Side highway would be underwater in 20 years. He did, however, predict that sea levels would rise by about 2 feet by 2100, which would put the West Side highway at risk of flooding. \\
\midrule
\textbf{Assistance} & The statement overlooks the importance of communicating uncertainty in climate science. As per the answer, Hansen's comment was speculative and not based on a specific study, thus it's crucial to mention this uncertainty to avoid misinterpretation of the potential impacts of climate change. & The statement fails to acknowledge that the answer does not explicitly convey the inherent uncertainty in climate predictions, such as Hansen's prediction about sea level rise. According to Wikipedia, "Uncertainty is a fundamental characteristic of weather, seasonal climate, and hydrological prediction, and no forecast is complete without a description of its uncertainty."  \\
\midrule
\textbf{Issues} &  & uncertainty\_missing \\
\midrule
\textbf{Rating} & 4 & 2.3 \\
\bottomrule
\end{tabular}
\caption{Example for the \textbf{uncertainty} dimension. The Model B answer states precise predictions without  acknowledging the uncertainty in climate science as there are many factors that can influence a prediction.}
\label{tab:example_uncertainty}
\end{table}

\subsection{AI Assistance}
In this section, we present examples showing AI assistance that influence human ratings in Tables \ref{tab:critique_tone} to \ref{tab:critique_uncertainty}.

\begin{table}[htp]
\centering
\small
\begin{tabular}{@{}p{4cm}p{9.5cm}@{}}
\toprule
\textbf{Question} & Is IPCC consensus phoney? \\
\midrule
\textbf{Answer} & No, the consensus of the Intergovernmental Panel on Climate Change (IPCC) is not phoney. The IPCC is a reputable international body that assesses the science related to climate change. It was established by the United Nations Environment Programme (UNEP) and the World Meteorological Organization (WMO) in 1988 to provide policymakers with regular scientific assessments on climate change, its implications and potential future risks, as well as to put forward adaptation and mitigation strategies. The consensus reached by the IPCC is based on rigorous scientific research and review by experts from all over the world, making it a reliable and authoritative source of information on climate change. \\
\midrule
\textbf{Assistance} & The tone of the answer, while factual, may be perceived as biased due to the strong defense of the IPCC's credibility and authority.  \\
\midrule
\textbf{Rating with Assistance} & 2.3 [biased, persuasive] \\
\textbf{Rating without Assistance} & 4.6 \\
\bottomrule
\end{tabular}
\caption{Example for the \textbf{tone} dimension. Raters identify the biased tone only with assistance.}
\label{tab:critique_tone}
\end{table}

\begin{table}[htp]
\centering
\small
\begin{tabular}{@{}p{4cm}p{9.5cm}@{}}
\toprule
\textbf{Question} & How do single-family houses contribute to higher toxic chemical emissions compared to multifamily buildings? \\
\midrule
\textbf{Answer} & Single-family houses contribute to higher toxic chemical emissions compared to multifamily buildings primarily due to their larger size and individualized systems. Each single-family home typically has its own heating, cooling, and water systems, all of which consume energy and produce emissions. Additionally, these houses often have larger lawns requiring maintenance equipment like lawnmowers and leaf blowers, which also emit harmful chemicals. Lastly, single-family homes, due to their larger size, require more materials to build and maintain, leading to higher emissions from the production and transportation of these materials. \\
\midrule
\textbf{Assistance} & While the answer provides a plausible explanation, it lacks empirical evidence or scientific findings to support the claim that single-family houses contribute to higher toxic chemical emissions compared to multifamily buildings.  \\
\midrule
\textbf{Rating with Assistance} & 1.6 [anecdotal, incorrect, science\_out\_of\_context] \\
\textbf{Rating without Assistance} & 4 \\
\bottomrule
\end{tabular}
\caption{Example for the \textbf{accuracy} dimension. Raters identify the anecdotal nature of the answer only with assistance.}
\label{tab:critique_accuracy}
\end{table}

\begin{table}[htp]
\centering
\small
\begin{tabular}{@{}p{4cm}p{9.5cm}@{}}
\toprule
\textbf{Question} & Are climate change solutions too expensive? \\
\midrule
\textbf{Answer} & While it's true that implementing solutions to combat climate change requires significant financial investment, it's important to consider the cost of inaction, which is likely to be much higher. The economic damage caused by extreme weather events, rising sea levels, and other climate-related impacts could be devastating. Moreover, many climate solutions, such as renewable energy technologies, are becoming increasingly cost-competitive. Investing in these solutions can also stimulate economic growth and create jobs, providing a return on investment. Therefore, while the upfront costs may be high, the long-term benefits both economically and environmentally make it a necessary investment. \\
\midrule
\textbf{Assistance} & The statement is incorrect as the answer contains a vague statement. The phrase "the cost of inaction, which is likely to be much higher" is not quantified or supported with specific data or examples. As per the provided Wikipedia paragraphs, "a 2021 study by the reinsurance company Swiss Re estimated global climate change is likely to reduce global economic output by 11-14\%, or as much as \$23 trillion annually by 2050, compared with output without climate change." This specific information could have been used to substantiate the claim.  \\
\midrule
\textbf{Rating with Assistance} & 2 [vague, irrelevant\_info] \\
\textbf{Rating without Assistance} & 4.6 \\
\bottomrule
\end{tabular}
\caption{Example for the \textbf{specificity} dimension. Raters identify how specific information could make the answer better only with assistance.}
\label{tab:critique_specificity}
\end{table}

\begin{table}[htp]
\centering
\small
\begin{tabular}{@{}p{4cm}p{9.5cm}@{}}
\toprule
\textbf{Question} & Is the Arctic ice melt a natural cycle? \\
\midrule
\textbf{Answer} & While the Arctic ice does go through natural cycles of melting and refreezing, the accelerated rate of ice melt that we are currently witnessing is not part of a natural cycle. Scientific evidence overwhelmingly points to human-induced climate change as the primary cause. Rising global temperatures, largely due to greenhouse gas emissions from human activities, are causing the Arctic ice to melt at an unprecedented rate. This rapid ice loss is having profound effects on Arctic ecosystems and contributing to rising sea levels globally. \\
\midrule
\textbf{Assistance} & While the answer provides a comprehensive explanation of the Arctic ice melt, it lacks specific details such as the decline rate of 4.7\% per decade, the projection of the first ice-free conditions around 2035 under a scenario of continually accelerating greenhouse gas emissions, and the fact that Arctic ice decline between 1979 and 2011 is estimated to have been responsible for as much radiative forcing as a quarter of CO2 emissions the same period.  \\
\midrule
\textbf{Rating with Assistance} & 2 [not\_enough\_detail] \\
\textbf{Rating without Assistance} & 5 \\
\bottomrule
\end{tabular}
\caption{Example for the \textbf{completeness} dimension. Raters identify how specific information could make the answer better only with assistance.}
\label{tab:critique_completeness}
\end{table}

\begin{table}[htp]
\centering
\small
\begin{tabular}{@{}p{4cm}p{9.5cm}@{}}
\toprule
\textbf{Question} & will global warming cause an ice age \\
\midrule
\textbf{Answer} & While it may seem counterintuitive, some scientists have suggested that global warming could potentially trigger what's known as a "mini ice age." This theory is based on the idea that increased melting of polar ice could disrupt ocean currents, such as the Gulf Stream, that play a crucial role in regulating global climate. If these currents were to be significantly altered or shut down, it could potentially lead to a dramatic cooling of certain regions, such as Northern Europe. However, this is just one theory and the overall consensus among scientists is that global warming will lead to a general trend of increasing temperatures worldwide. \\
\midrule
\textbf{Assistance} & While the answer acknowledges the theory of a potential "mini ice age" due to global warming, it fails to adequately convey the scientific consensus on the matter. As stated in the provided paragraphs, "Earth system models used in Coupled Model Intercomparison Project indicate that shutdown is only likely after high levels of warming are sustained well after 2100," and "paleoceanographic research suggests that the AMOC may be even more stable than what is predicted by most models." Therefore, the answer could mislead readers into thinking that a "mini ice age" is a more immediate and likely consequence of global warming than what is generally agreed upon by the scientific community.  \\
\midrule
\textbf{Rating with Assistance} & 1.6 [uncertainty\_missing, consensus\_missing, contradicting\_evidence\_missing] \\
\textbf{Rating without Assistance} & 4.6 \\
\bottomrule
\end{tabular}
\caption{Example for the \textbf{uncertainty} dimension. Raters identify the unlikelihood of the theory only with assistance.}
\label{tab:critique_uncertainty}
\end{table}

\end{document}